\documentclass{article} 
\usepackage{iclr2027_conference,times}

\usepackage{amsmath,amsfonts,bm}

\def\eqref#1{equation~\ref{#1}}

\def\1{\bm{1}}

\DeclareMathAlphabet{\mathsfit}{\encodingdefault}{\sfdefault}{m}{sl}
\SetMathAlphabet{\mathsfit}{bold}{\encodingdefault}{\sfdefault}{bx}{n}

\usepackage{hyperref}
\usepackage{url}
\usepackage{booktabs}
\usepackage{graphicx}
\usepackage{amssymb}
\usepackage{multirow}
\usepackage{bbm}
\usepackage{caption}
\usepackage{wrapfig}
\usepackage{xcolor}

\usepackage{xcolor}
\usepackage{booktabs}
\usepackage{graphicx}
\definecolor{bestgreen}{RGB}{220,240,220}
\definecolor{secondorange}{RGB}{255,235,210}

\newcommand{\sotabox}[1]{%
    \begingroup
    \setlength{\fboxsep}{1.5pt}%
    \colorbox{bestgreen}{\strut\textbf{#1}}%
    \endgroup
}

\newcommand{\secondbox}[1]{%
    \begingroup
    \setlength{\fboxsep}{1.5pt}%
    \colorbox{secondorange}{\strut\underline{#1}}%
    \endgroup
}

\usepackage{xspace}

\usepackage[capitalize]{cleveref}
\crefname{section}{Sec.}{Secs.}
\Crefname{section}{Section}{Sections}
\Crefname{table}{Table}{Tables}
\crefname{table}{Tab.}{Tabs.}

\makeatletter
\DeclareRobustCommand\onedot{\futurelet\@let@token\@onedot}
\def\@onedot{\ifx\@let@token.\else.\null\fi\xspace}

\let\@authorsaddresses\@empty
\makeatother

\usepackage{xcolor}
\usepackage{pifont}
\usepackage{amsmath}
\newcommand{\cmark}{\textcolor{green!60!black}{\checkmark}}
\newcommand{\xmark}{\textcolor{red!70!black}{\ding{55}}}

\newcommand{\ours}{F\textsuperscript{2}SLAM}
\newcommand{\PAR}[1]{\vspace{0.1cm}\noindent{\bf #1} }

\title{\textbf{\ours{}}: Turning Feed-Forward Geometry into Persistent Factors for SLAM}

\author{Zhisong Xu$^{1\dag}$ \hspace{0.61em}
Fan Zhu$^{2\dag}$ \hspace{0.61em}
Jiawei Qian$^{1}$ \hspace{0.61em}
Ziyu Chen$^{3}$ \hspace{0.61em}
Zhenjun Zhao$^{4}$\thanks{Corresponding author. $^{\dag}$Equal contribution.} \hspace{0.61em}
Javier Civera$^{4}$ \\
$^{1}$The University of Tokyo
\quad
$^{2}$Tohoku University\\
$^{3}$University of Science and Technology of China
\quad
$^{4}$University of Zaragoza
}
\iclrfinalcopy 
\begin{document}

\maketitle

\begin{abstract}
Feed-forward 3D models provide strong multi-view geometric priors, while online simultaneous localization and mapping (SLAM) relies mainly on local measurements and can accumulate drift over long sequences.
Existing attempts to combine the two typically treat feed-forward predictions as an external geometric state that is aligned or fused with the online estimate after the fact, which keeps broader multi-view evidence outside the optimizer that refines the SLAM state.
We present \textbf{\ours{}}, which instead converts feed-forward geometry directly into optimization-native target-weight measurements attached to a persistent dense factor graph.
A high-frequency stream maintains local tracking constraints and graph connectivity, while a low-frequency stream uses wider multi-view context to selectively refresh existing measurements after a state-consistency check.
Both streams constrain the same poses, inverse depths, and optional camera intrinsics through a single dense bundle adjustment.
Experiments on multiple benchmarks demonstrate consistently strong trajectory estimation and improved dense reconstruction in both calibrated and uncalibrated settings.
Notably, the uncalibrated configuration reduces the average ATE RMSE from $0.030$\,m for the strongest feed-forward baseline to $0.002$\,m on the Replica dataset.
\end{abstract}

\section{Introduction}

Online visual simultaneous localization and mapping (SLAM) estimates camera motion and scene geometry by maintaining an optimizable geometric state that dense bundle adjustment (BA) continuously revisits, re-weights, and jointly refines as new observations arrive~\citep{orbslam2,droidslam,zhao2026advances}.
Because these measurements are drawn mainly from short temporal windows, online SLAM remains sensitive to weak overlap, occlusion, viewpoint changes, and accumulated tracking error over long sequences.

Recent feed-forward 3D foundation models (FMs)~\citep{vggt} offer a complementary form of geometric reasoning.
By jointly processing multiple images, they directly infer depth, point maps, camera poses, and multi-view structure over much broader context than a local BA window~\citep{PanoImager}.
This has motivated foundation-model-based SLAM (FM-SLAM) systems such as MASt3R-SLAM~\citep{mast3rslam}, VGGT-SLAM~\citep{vggtslam}, and VGGT-SLAM++~\citep{vggtslam++}, which exploit learned geometry for tracking, mapping, relocalization, loop closure, or submap alignment.

\begin{figure*}[!t]
    \centering
    \includegraphics[width=0.98\linewidth]{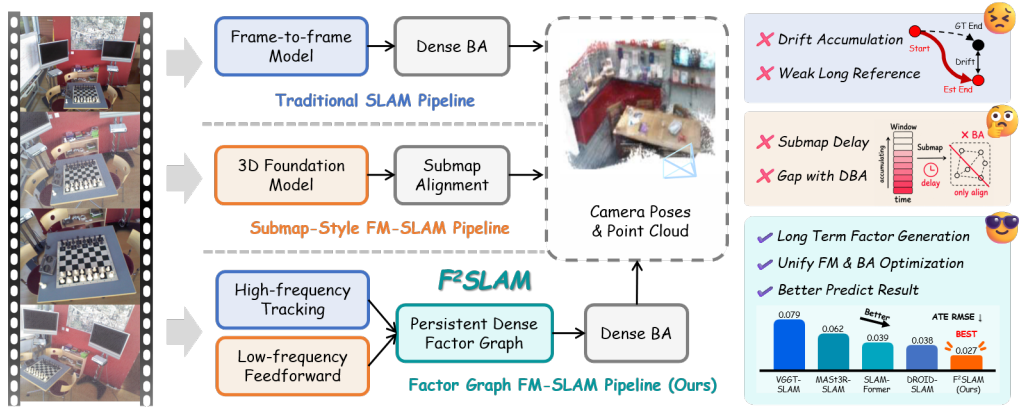}
    \caption{
        Motivation and overview of \protect\ours{}. Conventional dense SLAM (top) relies on local frame-to-frame measurements, while foundation-model-based approaches (middle) reconstruct and align local submaps. In contrast, \protect\ours{} (bottom) integrates high-frequency tracking and low-frequency feed-forward geometry in a persistent dense factor graph, where low-frequency predictions selectively refresh existing measurements for joint refinement through a single dense BA. The inset reports ATE RMSE on the TUM RGB-D dataset~\cite{tum}.
    }
    \label{fig:intro}
\end{figure*}

These systems, however, typically compute feed-forward predictions over local windows or submaps and only afterwards reconcile them with the SLAM state through alignment, fusion, or state-level correction, as shown in \cref{fig:intro}.
Feed-forward geometry therefore remains an external state rather than evidence that the online optimizer can act on directly.
This separation is costly in two ways. 
Broader multi-view evidence never enters the measurement set that dense BA actually optimizes.
Once a submap has been aligned, it also cannot be jointly re-weighted against subsequent local evidence, since the correction is applied only once.
The missing link is a way to make feed-forward geometric evidence directly and repeatedly participate in persistent SLAM optimization, rather than serve as a one-time external correction.



We address this with \textbf{\ours{}}, which follows one guiding principle: \emph{feed-forward geometry should enter SLAM as optimization-native measurements, not as a separate geometric state.}
\ours{} converts multi-view feed-forward evidence into dense target-weight measurements attached directly to the existing factor graph, so that feed-forward reasoning and recurrent tracking constrain the same keyframe poses, dense inverse depths, and optional camera intrinsics through one dense BA backend.
Doing so online requires updating these measurements at two different timescales.
A high-frequency stream maintains local constraints and graph connectivity, while a low-frequency stream uses wider multi-view context to selectively refresh existing measurements.
Because feed-forward predictions can be unreliable under occlusion or weak overlap, the central difficulty is deciding when a refresh should be trusted. \ours{} addresses this with a state-consistency check that admits a candidate measurement only when it agrees with the current geometric state, and otherwise leaves the existing measurement unchanged.
Feed-forward pose and depth predictions may also provide gauge-aligned initialization, but subsequent geometric refinement remains entirely within the persistent BA state.

In summary, our \textbf{contributions} are as follows:
\begin{itemize}

\item
We formulate feed-forward multi-view geometry as optimization-native target-weight measurements that directly participate in persistent dense BA.

\item
We develop a dual-timescale measurement-refresh mechanism with state-consistent admission, enabling local tracking and broader multi-view reasoning to update a common factor representation within a single optimized geometric state.


\item
We instantiate this formulation in a complete online system, \ours{}, and demonstrate strong trajectory estimation and dense reconstruction across calibrated and uncalibrated indoor and outdoor benchmarks.

\end{itemize}
\section{Related Work}


\PAR{Dense RGB SLAM.}
Dense RGB SLAM estimates camera motion and scene geometry from video with photometric, geometric, or learned constraints.
Classical methods such as ORB-SLAM~\citep{orbslam2,orbslam3} and LSD-SLAM~\citep{lsdslam} rely on hand-crafted features, direct alignment, and bundle adjustment, while learning-based methods~\citep{droidslam,dpvslam} integrate learned depth, matching, confidence, and motion components for robust estimation.
Recent NeRF- and 3DGS-based methods~\citep{nicerslam, goslam, mygosplat, FGO-SLAM++} improve scene representation but still struggle with long-range consistency~\citep{LR-SGS}.


\PAR{Feed-forward 3D reconstruction priors.}
Feed-forward reconstruction models have become strong priors for multi-view geometry.
Pairwise methods such as DUSt3R~\citep{dust3r} and MASt3R~\citep{mast3r} predict dense structure and correspondences for wide-baseline reconstruction, while multi-view models such as VGGT~\citep{vggt} extend this to joint camera, depth, point-map, and track prediction.
Streaming variants~\citep{zhuo2025streaming,lingbotmap, framevggt} support online or long-sequence reconstruction with memory or causal processing, but remain window-dependent and lack persistent sequence-level optimization.


\PAR{SLAM with feed-forward reconstruction priors.}
Recent foundation-model-based SLAM systems~\citep{slam3r,mast3rslam,vggtslam} have explored using feed-forward reconstruction priors for online tracking and mapping, often by applying them to local windows or submaps and then aligning, fusing, or correcting the resulting poses, depths, and point maps.
However, the correction they provide is applied only once, since feed-forward predictions either sit outside the main dense BA or are optimized through a separate alignment stage that is not revisited as new local evidence arrives.
\ours{} instead converts feed-forward outputs into dense target-weight factors and jointly optimizes them with local DROID-style factors in one dense BA backend, directly coupling online tracking with feed-forward measurements predicted from broader multi-view context.
\section{Feed-forward Factor SLAM}
\label{sec:method}

\begin{figure}[t]
    \centering
    \includegraphics[width=0.98\linewidth]{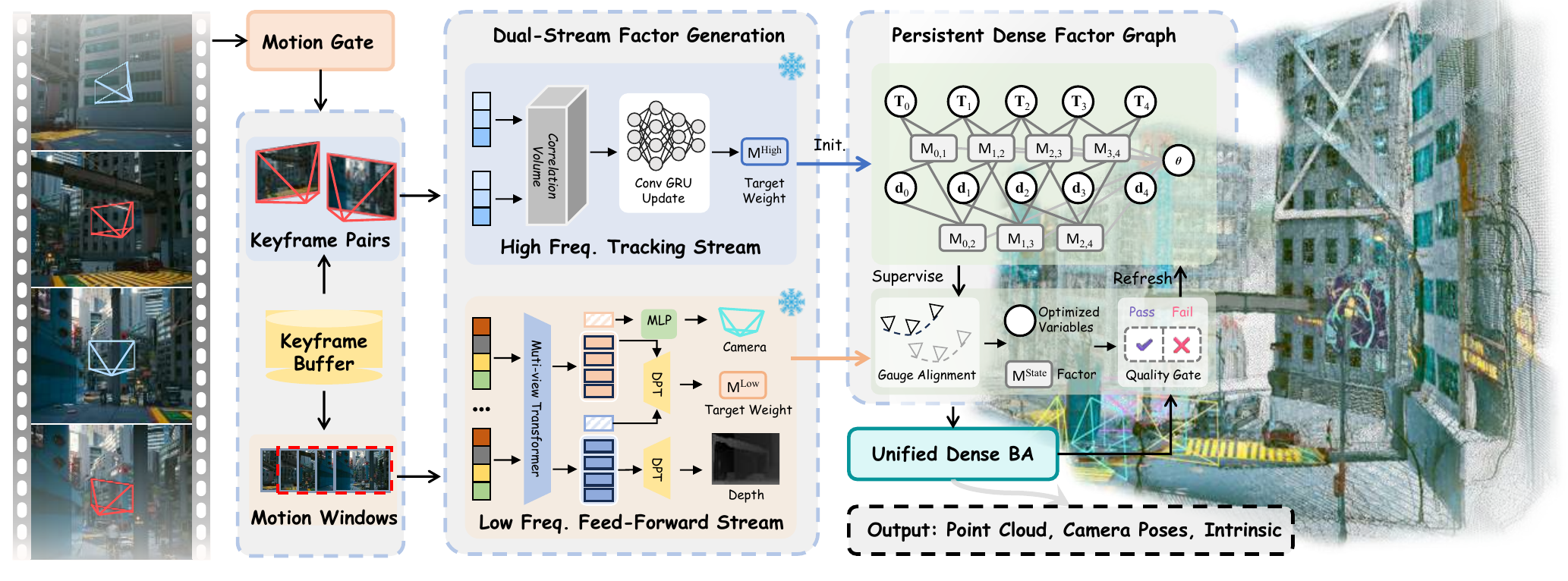}
    \caption{
    Pipeline of \ours{}.
    The high-frequency stream selects keyframes, maintains the graph topology, and frequently updates local target-weight measurements.
    The low-frequency stream refreshes measurements on selected existing edges using wider multi-view context, without adding new edges.
    Both streams constrain the same persistent geometric state through a single dense BA.
    }
    \label{fig:pipeline}
\end{figure}

\subsection{Persistent Factor-Graph Formulation}
\label{sec:formulation}

Existing feed-forward SLAM systems typically treat feed-forward geometry as an external state that must be aligned or fused with the online estimate.
In contrast, \ours{} converts feed-forward predictions into dense measurements that directly constrain the same persistent geometric state as online tracking.
\cref{fig:pipeline} gives the pipeline of \ours{}.

Given an RGB stream $\{I_t\}_{t=1}^{T}$, we maintain an ordered set of keyframe indices:
\begin{equation}
    \mathcal{K}_t
    =
    \{k_1,\ldots,k_{N_t}\},
    \label{eq:keyframe_set}
\end{equation}
where $k_{N_t}$ is the latest keyframe.
The persistent geometric state is:
\begin{equation}
    \mathcal{X}_t
    =
    \left\{\mathbf{T}_i,\mathbf{d}_i\right\}_{i\in\mathcal{K}_t}
    \cup
    \left\{\boldsymbol{\theta}\right\},
    \label{eq:state}
\end{equation}
where $\mathbf{T}_i\in SE(3)$ is the world-to-camera transformation, $\mathbf{d}_i\in\mathbb{R}_{+}^{h\times w}$ is the dense inverse-depth map, and $\boldsymbol{\theta}$ denotes the camera intrinsics.
When calibration is known, $\boldsymbol{\theta}$ is fixed and omitted from the optimized variables.
We denote the current estimate by $\bar{\mathcal{X}}_t$ and the result of a BA solver by $\mathcal{X}_t^{*}$.

The active portion of the persistent factor graph is:
\begin{equation}
    \mathcal{G}_t
    =
    \left(
        \mathcal{K}_t,
        \mathcal{E}_{\mathrm{act}}^t
    \right),
    \label{eq:factor_graph}
\end{equation}
where $\mathcal{E}_{\mathrm{act}}^t$ contains the active directed edges.
Each edge $(i,j)$ stores a dense target-weight measurement:
\begin{equation}
    \mathcal{M}_{ij}
    =
    \left(
        \mathbf{p}^{*}_{ij},
        \mathbf{w}_{ij}
    \right),
    \label{eq:measurement}
\end{equation}
where $\mathbf{p}^{*}_{ij}(\mathbf{p})\in\mathbb{R}^{2}$ is the predicted target coordinate of source pixel $\mathbf{p}\in\Omega_i$, and $\mathbf{w}_{ij}(\mathbf{p})\in\mathbb{R}_{+}^{2}$ contains the confidence weights for the two reprojection-residual components.
We define:
\begin{equation}
    \mathbf{\Lambda}_{ij}(\mathbf{p})
    =
    \operatorname{diag}\!\left(\mathbf{w}_{ij}(\mathbf{p})\right),
    \label{eq:confidence_matrix}
\end{equation}
which weights the dense reprojection residual in the BA objective.

All measurements in $\mathcal{G}_t$ use this target-weight representation.
Feed-forward predictions may initialize the shared geometric state or generate factor measurements, but do not introduce a separate optimized state.
Only $\mathcal{X}_t$ is optimized by the common dense BA.

\subsection{Dual-stream Pair Construction and Factor Generation}
\label{sec:factor_generation}
We update the shared factor representation through two streams operating at different timescales.
Following standard visual SLAM practice~\citep{orbslam2}, incoming frames are selected as keyframes according to their image-space motion relative to the latest keyframe.
Detailed insertion and low-frequency scheduling rules are provided in Appendix~\ref{app:motion_keyframe}.

After inserting a new keyframe $k_{N_t}$, its temporal and geometrically proximal neighbors define:
\begin{equation}
    \mathcal{N}_{\mathrm{high}}(k_{N_t})
    =
    \mathcal{N}_{\mathrm{temp}}(k_{N_t})
    \cup
    \mathcal{N}_{\mathrm{prox}}(k_{N_t}),
    \label{eq:high_neighbors}
\end{equation}
from which the high-frequency stream creates directed pairs:
\begin{equation}
    \mathcal{P}_{\mathrm{high}}^t
    =
    \bigcup_{j\in\mathcal{N}_{\mathrm{high}}(k_{N_t})}
    \left\{(k_{N_t},j),(j,k_{N_t})\right\}.
    \label{eq:high_pairs}
\end{equation}
These pairs are inserted into $\mathcal{E}_{\mathrm{act}}^t$, while factors exceeding the update-age or capacity limits are retired with their measurements retained for possible reuse in local BA.
Thus, the high-frequency stream constructs and maintains the persistent graph topology.

Each active edge stores:
\begin{equation}
    \mathcal{F}_{ij}
    =
    \left(
        \mathcal{M}_{ij},
        \ell_{ij},
        \gamma_{ij}
    \right),
    \label{eq:factor_state}
\end{equation}
where $\ell_{ij}\in\{\mathrm{high},\mathrm{low}\}$ records the measurement source and $\gamma_{ij}\in\{0,1\}$ controls whether high-frequency updates may overwrite it.
Note that these metadata are not optimizable geometric variables.

For each unlocked edge $(i,j)\in\mathcal{E}_{\mathrm{act}}^t$, the high-frequency recurrent updater predicts:
\begin{equation}
\begin{aligned}
    \big(
        \mathbf{h}_{ij}^{+},
        \Delta\mathbf{p}^{\mathrm{high}}_{ij},
        \mathbf{w}^{\mathrm{high}}_{ij}
    \big)
    =
    \Psi_{\mathrm{high}}
    \big(
        \mathbf{h}_{ij},
        \mathbf{x}_i,
        \mathbf{C}_{ij},
        \mathbf{m}_{ij}
    \big).
\end{aligned}
\label{eq:high_prediction}
\end{equation}
Here, $\mathbf{h}_{ij}$ and $\mathbf{h}_{ij}^{+}$ are the recurrent hidden states, $\mathbf{x}_i$ is the source context feature, $\mathbf{C}_{ij}$ contains correlation features sampled at the current reprojection, and $\mathbf{m}_{ij}$ encodes geometry-induced flow and the discrepancy from the stored target.
The resulting measurement is:
\begin{equation}
\begin{aligned}
    \mathbf{p}^{*,\mathrm{high}}_{ij}(\mathbf{p})
    &=
    \hat{\mathbf{p}}_{ij}
    \left(\mathbf{p};\bar{\mathcal{X}}_t\right)
    +
    \Delta\mathbf{p}^{\mathrm{high}}_{ij}(\mathbf{p}),\\
    \mathcal{F}_{ij}
    &\leftarrow
    \left(
        \left(
            \mathbf{p}^{*,\mathrm{high}}_{ij},
            \mathbf{w}^{\mathrm{high}}_{ij}
        \right),
        \mathrm{high},
        0
    \right).
\end{aligned}
\label{eq:high_factor_update}
\end{equation}
The constructed target remains fixed during the ensuing BA solve.

Meanwhile, newly inserted keyframes are accumulated into a scheduled low-frequency window:
\begin{equation}
    \mathcal{W}_t
    =
    \{k_{a_t},\ldots,k_{b_t}\},
    \label{eq:low_window}
\end{equation}
which provides wider multi-view context rather than defining a separate submap or factor graph.
The low-frequency stream operates only on selected existing active edges:
\begin{equation}
    \mathcal{P}_{\mathrm{low}}^t
    \subseteq
    \left\{
        (i,j)\in\mathcal{E}_{\mathrm{act}}^t
        \;\middle|\;
        i,j\in\mathcal{W}_t,\ i\neq j
    \right\}.
    \label{eq:low_pairs}
\end{equation}

Multi-view features are extracted as:
\begin{equation}
    \{\mathbf{H}_i\}_{i\in\mathcal{W}_t}
    =
    \mathrm{MVTrans}
    \left(
        \{E_{\mathrm{feat}}(I_i)\}_{i\in\mathcal{W}_t}
    \right),
    \label{eq:low_features}
\end{equation}
where $\mathrm{MVTrans}$ aggregates per-image backbone features across the window.
When enabled, gauge-aligned pose and depth predictions initialize the shared geometric state before pair selection and validation, but do not directly modify the predicted flow targets.

For each $(i,j)\in\mathcal{P}_{\mathrm{low}}^t$, the multi-view factor decoder $H_{\mathrm{factor}}$ converts the multi-view features into a dense displacement-confidence pair:
\begin{equation}
    \left(
        \mathbf{f}^{\mathrm{low}}_{ij},
        \mathbf{c}^{\mathrm{low}}_{ij}
    \right)
    =
    H_{\mathrm{factor}}
    \left(
        \mathbf{H}_i,\mathbf{H}_j
    \right),
    \label{eq:low_factor_prediction}
\end{equation}
where $H_{\mathrm{factor}}$ includes decoding and conversion to the common BA grid.
Importantly, $\mathbf{f}^{\mathrm{low}}_{ij}$ is the full source-to-target displacement in BA coordinates, rather than a correction to the current geometric reprojection.
The confidence vector $\mathbf{c}^{\mathrm{low}}_{ij}$ uses identical horizontal and vertical components derived from the scalar confidence gate.

The candidate measurement is constructed for each source pixel $\mathbf{p}\in\Omega_i$ as:
\begin{equation}
\begin{aligned}
    \mathbf{p}^{*,\mathrm{low}}_{ij}(\mathbf{p})
    &=
    \mathbf{p}
    +
    \mathbf{f}^{\mathrm{low}}_{ij}(\mathbf{p}),\\
    \mathbf{w}^{\mathrm{low}}_{ij}
    &=
    \eta_{\mathrm{low}}
    \mathbf{c}^{\mathrm{low}}_{ij},
\end{aligned}
\label{eq:low_target_weight}
\end{equation}
where $\eta_{\mathrm{low}}>0$ is a fixed scalar controlling the relative weight of low-frequency measurements.
The resulting candidate measurement is denoted by:
\begin{equation}
    \mathcal{M}^{\mathrm{low}}_{ij}
    =
    \left(
        \mathbf{p}^{*,\mathrm{low}}_{ij},
        \mathbf{w}^{\mathrm{low}}_{ij}
    \right).
    \label{eq:low_candidate_measurement}
\end{equation}
These candidates are validated against the current SLAM state before being allowed to refresh the corresponding existing factor measurements.

\subsection{State-Consistent Low-Frequency Factor Refresh}
\label{sec:graph_update}

Low-frequency predictions may be unreliable under occlusion or limited overlap.
We therefore validate each candidate against the current SLAM state before refreshing its existing factor measurement.
For $(i,j)\in\mathcal{P}_{\mathrm{low}}^t$, the reprojection residual is:
\begin{equation}
    \mathbf{e}_{ij}^{\mathrm{low}}(\mathbf{p})
    =
    \hat{\mathbf{p}}_{ij}
    \left(\mathbf{p};\bar{\mathcal{X}}_t\right)
    -
    \mathbf{p}_{ij}^{*,\mathrm{low}}(\mathbf{p}),
    \label{eq:low_candidate_residual}
\end{equation}
where $\hat{\mathbf{p}}_{ij}(\mathbf{p};\bar{\mathcal{X}}_t)$ denotes the reprojection of source pixel $\mathbf{p}$ into keyframe $j$ under the current pose and inverse-depth estimates $\bar{\mathcal{X}}_t$.


Validating a candidate requires aggregating this residual, together with its predicted confidence, into per-pair statistics.
A source pixel belongs to the \emph{support} set when its projection is valid and its confidence exceeds a fixed floor, and is further counted as an \emph{inlier} when its reprojection residual additionally satisfies $\|\mathbf{e}_{ij}^{\mathrm{low}}(\mathbf{p})\|_2 \leq \tau_{\mathrm{geo}}$, where $\tau_{\mathrm{geo}}$ is a fixed pixel-level tolerance.
We then compute an acceptance score for each candidate pair:
\begin{equation}
    s_{ij}
    =
    \frac{
        \bar{c}_{ij}r_{\mathrm{in}}^{ij}
        \sqrt{r_{\mathrm{sup}}^{ij}}
    }{
        1+\bar{e}_{ij}^{\mathrm{low}}
    },
    \label{eq:low_acceptance_score}
\end{equation}
where $\bar{c}_{ij}$ denotes the mean confidence over the support set, $r_{\mathrm{sup}}^{ij}$ and $r_{\mathrm{in}}^{ij}$ the support and inlier ratios defined above, and $\bar{e}_{ij}^{\mathrm{low}}$ the mean reprojection error over inliers.

A candidate is accepted when:
\begin{equation}
    (i,j)\in\mathcal{S}_{\mathrm{low}}^t
    \iff
    r_{\mathrm{sup}}^{ij}\geq\tau_{\mathrm{sup}}
    \;\land\;
    s_{ij}\geq\tau_{\mathrm{score}}.
    \label{eq:accepted_low_edges}
\end{equation}
where $\mathcal{S}_{\mathrm{low}}^t$ denotes the set of accepted low-frequency pairs, and $\tau_{\mathrm{sup}}$ and $\tau_{\mathrm{score}}$ are fixed thresholds for support and the acceptance score, respectively.
Accepted candidates refresh the existing factor-state as:
\begin{equation}
    \mathcal{F}_{ij}
    \leftarrow
    \left(
        \mathcal{M}_{ij}^{\mathrm{low}},
        \mathrm{low},
        1
    \right),
    \qquad
    (i,j)\in\mathcal{S}_{\mathrm{low}}^t,
    \label{eq:low_factor_state_update}
\end{equation}
and remain protected from high-frequency updates for the remaining factor lifetime.
Rejected candidates leave the factor unchanged.
Thus, low-frequency refresh changes only the stored measurement of an existing edge, without altering the graph topology or introducing a separate geometric state.

The two streams differ substantially in architecture, a local recurrent correlation-volume updater versus a global multi-view transformer, yet both are required to produce a candidate in the same BA coordinate system and the same target-weight format.
Gauge alignment removes cross-source scale and reference-frame discrepancies before a candidate is even considered, and the acceptance criterion in \cref{eq:accepted_low_edges} evaluates every candidate against the current geometric state rather than trusting either source's own confidence estimate.
Unification is therefore enforced by this admission procedure, not merely asserted by the shared measurement format.

\subsection{Dense Bundle Adjustment}
\label{sec:dense_ba}

\ours{} jointly optimizes the persistent geometric state using the measurements selected for the current BA solve.
The backend is agnostic to their source.
$\ell_{ij}$ and $\gamma_{ij}$ affect measurement updates but do not enter the objective.

For a source pixel $\mathbf{p}\in\Omega_i$, its projection into keyframe $j$ is:
\begin{equation}
    \hat{\mathbf{p}}_{ij}
    \left(\mathbf{p};\mathcal{X}_t\right)
    =
    \pi_{\boldsymbol{\theta}}
    \left(
        \mathbf{T}_{j}\mathbf{T}_{i}^{-1}
        \,
        \pi_{\boldsymbol{\theta}}^{-1}
        \left(\mathbf{p},\mathbf{d}_i(\mathbf{p})\right)
    \right),
    \label{eq:dense_ba_projection}
\end{equation}
yielding the reprojection residual:
\begin{equation}
    \mathbf{r}_{ij}
    \left(\mathbf{p};\mathcal{X}_t\right)
    =
    \hat{\mathbf{p}}_{ij}
    \left(\mathbf{p};\mathcal{X}_t\right)
    -
    \mathbf{p}^{*}_{ij}(\mathbf{p}).
    \label{eq:dense_ba_residual}
\end{equation}
Here, $\mathcal{E}_{\mathrm{BA}}^t$ contains the active factors and any recent inactive factors selected for the current solve.

Using $\mathbf{\Lambda}_{ij}$ from \cref{eq:confidence_matrix}, we optimize:
\begin{equation}
\begin{aligned}
    \mathcal{X}_t^{*}
    =
    \arg\min_{\mathcal{X}_t}\,
    &\sum_{(i,j)\in\mathcal{E}_{\mathrm{BA}}^t}
    \sum_{\mathbf{p}\in\Omega_i}
    v_{ij}(\mathbf{p})\,
    \rho\!\left(
        \mathbf{r}_{ij}(\mathbf{p};\mathcal{X}_t)^{\top}
        \mathbf{\Lambda}_{ij}(\mathbf{p})
        \mathbf{r}_{ij}(\mathbf{p};\mathcal{X}_t)
    \right)\\
    &+
    \frac{\lambda_d}{2}
    \sum_{i\in\mathcal{K}_t}
    \|\nabla\mathbf{d}_i\|_2^2
    +
    \frac{\lambda_{\theta}}{2}
    \|\boldsymbol{\theta}-\boldsymbol{\theta}_0\|_{\mathbf{\Sigma}_{\theta}^{-1}}^2,
\end{aligned}
\label{eq:dense_ba_objective}
\end{equation}
where $v_{ij}$ masks invalid correspondences, $\rho$ is a robust penalty, the second term regularizes dense inverse depth, and the third term constrains camera intrinsics around the calibration prior $\boldsymbol{\theta}_0$.
The intrinsic term is omitted when calibration is fixed.

Each edge contributes a single residual using its currently stored measurement.
Thus, an accepted low-frequency prediction replaces the measurement on an existing edge without adding another residual or separate geometric state.
Both streams constrain the same persistent state through a single dense BA.
Detailed linearization and state updates are provided in Appendix~\ref{app:state_optimization}.
\section{Experiments}
\subsection{Experimental Setup}

\begin{table*}[t]
\centering
\small
\caption{
Trajectory evaluation (ATE RMSE [m] $\downarrow$) on the
TUM RGB-D dataset~\citep{tum}.
$*$ denotes results reproduced on our hardware.
}
\label{tab:tum_traj}

\resizebox{\textwidth}{!}{
\begin{tabular}{l|c|ccccccccc|c}
\toprule
Method & Uncalib.
& 360 & desk & desk2 & floor & plant
& room & rpy & teddy & xyz & Avg. \\
\midrule

DROID-SLAM~\citep{droidslam}
& \xmark
& 0.111
& 0.018
& 0.042
& \sotabox{0.021}
& \secondbox{0.016}
& \sotabox{0.049}
& 0.026
& 0.048
& 0.012
& 0.038 \\

DPV-SLAM~\citep{dpvslam}
& \xmark
& 0.112
& 0.018
& 0.029
& 0.057
& 0.021
& 0.330
& 0.030
& 0.084
& \secondbox{0.010}
& 0.076 \\

DPV-SLAM++~\citep{dpvslam}
& \xmark
& 0.132
& 0.018
& 0.029
& 0.050
& 0.022
& 0.096
& 0.032
& 0.098
& \secondbox{0.010}
& 0.054 \\

GO-SLAM~\citep{goslam}
& \xmark
& 0.089
& \sotabox{0.016}
& 0.028
& \secondbox{0.025}
& 0.026
& \secondbox{0.052}
& \sotabox{0.019}
& 0.048
& \secondbox{0.010}
& 0.035 \\

MASt3R-SLAM~\citep{mast3rslam}
& \xmark
& \sotabox{0.049}
& \sotabox{0.016}
& \sotabox{0.024}
& \secondbox{0.025}
& 0.020
& 0.061
& 0.027
& \secondbox{0.041}
& \sotabox{0.009}
& \secondbox{0.030} \\

\textbf{\ours{} (ours)}
& \xmark
& \secondbox{0.070}
& \secondbox{0.017}
& \secondbox{0.026}
& \sotabox{0.021}
& \sotabox{0.014}
& \sotabox{0.049}
& \secondbox{0.023}
& \sotabox{0.034}
& \secondbox{0.010}
& \sotabox{0.029} \\

\midrule

DROID-SLAM~\citep{droidslam}
& \cmark
& 0.202
& 0.032
& 0.091
& 0.064
& 0.045
& 0.918
& 0.056
& 0.045
& \secondbox{0.012}
& 0.163 \\

MASt3R-SLAM~\citep{mast3rslam}
& \cmark
& 0.070
& 0.035
& 0.055
& 0.056
& 0.035
& 0.118
& 0.041
& 0.114
& 0.020
& 0.060 \\

VGGT-SLAM~\citep{vggtslam}
& \cmark
& 0.071
& 0.025
& 0.040
& 0.141
& 0.023
& 0.102
& 0.030
& 0.034
& 0.014
& 0.053 \\


VGGT-SLAM 2.0$^{*}$~\citep{vggtslam2}
& \cmark
& \sotabox{0.050}
& 0.025
& 0.029
& 0.102
& 0.026
& \secondbox{0.063}
& 0.026
& 0.038
& 0.014
& 0.041 \\

SLAM-Former$^{*}$~\citep{SLAM-Former}
& \cmark
& 0.067
& \secondbox{0.018}
& \secondbox{0.026}
& 0.079
& \secondbox{0.021}
& 0.082
& \sotabox{0.017}
& \secondbox{0.030}
& \sotabox{0.011}
& \secondbox{0.039} \\

Lingbot-Map$^{*}$~\citep{lingbotmap}
& \cmark
& 0.091
& 0.034
& 0.038
& \secondbox{0.055}
& 0.032
& 0.064
& \secondbox{0.023}
& 0.039
& 0.016
& 0.044 \\

\textbf{\ours{} (ours)}
& \cmark
& \secondbox{0.066}
& \sotabox{0.016}
& \sotabox{0.023}
& \sotabox{0.015}
& \sotabox{0.013}
& \sotabox{0.043}
& 0.026
& \sotabox{0.029}
& \secondbox{0.012}
& \sotabox{0.027} \\

\bottomrule
\end{tabular}
}
\end{table*}
\begin{table*}[t]
\centering
\small
\caption{
Trajectory evaluation (ATE RMSE [m] $\downarrow$) on the
7-Scenes dataset~\citep{shotton2013scene}.
$*$ denotes results reproduced on our hardware.
}
\label{tab:7scenes_traj}

\resizebox{\textwidth}{!}{
\begin{tabular}{l|c|ccccccc|c}
\toprule
Method & Uncalib.
& chess & fire & heads & office
& pumpkin & kitchen & stairs & Avg. \\
\midrule

DROID-SLAM~\citep{droidslam}
& \xmark
& \secondbox{0.036}
& \sotabox{0.027}
& \sotabox{0.025}
& \sotabox{0.066}
& \sotabox{0.127}
& \secondbox{0.040}
& \secondbox{0.026}
& \sotabox{0.050} \\

NICER-SLAM~\citep{nicerslam}
& \xmark
& \sotabox{0.033}
& 0.069
& \secondbox{0.042}
& 0.108
& 0.200
& \sotabox{0.039}
& 0.108
& 0.086 \\

\textbf{\ours{} (ours)}
& \xmark
& \secondbox{0.036}
& \secondbox{0.029}
& \sotabox{0.025}
& \secondbox{0.088}
& \secondbox{0.140}
& 0.042
& \sotabox{0.021}
& \secondbox{0.053} \\

\midrule

DROID-SLAM~\citep{droidslam}
& \cmark
& 0.047
& 0.038
& 0.034
& 0.136
& 0.166
& 0.080
& 0.044
& 0.078 \\

MASt3R-SLAM~\citep{mast3rslam}
& \cmark
& 0.063
& 0.046
& 0.029
& 0.103
& \secondbox{0.114}
& 0.074
& 0.032
& 0.066 \\

VGGT-SLAM$^{*}$~\citep{vggtslam}
& \cmark
& 0.037
& \secondbox{0.026}
& \secondbox{0.018}
& 0.104
& 0.133
& 0.061
& 0.093
& 0.067 \\


VGGT-SLAM 2.0$^{*}$~\cite{vggtslam2}
& \cmark
& 0.039
& 0.027
& 0.021
& 0.106
& 0.135
& 0.053
& 0.092
& 0.067 \\

SLAM-Former$^{*}$~\cite{SLAM-Former}
& \cmark
& 0.040
& 0.033
& 0.020
& \sotabox{0.065}
& \sotabox{0.066}
& \sotabox{0.035}
& \secondbox{0.027}
& \sotabox{0.041} \\

SLAM3R~\cite{slam3r}
& \cmark
& 0.098
& 0.060
& 0.039
& \secondbox{0.078}
& 0.152
& 0.136
& 0.107
& 0.096 \\

Lingbot-Map$^{*}$~\citep{lingbotmap}
& \cmark
& \sotabox{0.034}
& 0.041
& 0.022
& 0.134
& 0.151
& 0.057
& 0.090
& 0.076 \\

\textbf{\ours{} (Ours)}
& \cmark
& \secondbox{0.036}
& \sotabox{0.024}
& \sotabox{0.014}
& 0.087
& 0.140
& \secondbox{0.041}
& \sotabox{0.020}
& \secondbox{0.051} \\

\bottomrule
\end{tabular}
}
\end{table*}

We evaluate \ours{} on four standard SLAM benchmarks: TUM RGB-D~\citep{tum}, 7-Scenes~\citep{shotton2013scene}, Replica~\citep{replica}, and KITTI Odometry~\citep{kitti} datasets.
Following~\citep{vggtslam}, we use the \texttt{evo} toolkit to Sim(3)-align estimated trajectories and report the root mean square error of absolute trajectory error (ATE RMSE).
Following~\citep{SLAM-Former}, dense reconstruction is evaluated using accuracy (Acc.) and completion (Comp.) on Replica, and additionally Chamfer distance (Chamfer) on 7-Scenes.
Unless noted otherwise, the \sotabox{best} and \secondbox{second-best} results in each column are highlighted in green (bold) and orange (underlined), respectively.
For benchmarks reported separately by calibration setting (\cref{tab:tum_traj,tab:7scenes_traj,tab:replica_traj,tab:kitti_traj}), rankings are computed within each setting.

For trajectory evaluation, we compare against a broad range of classical, learning-based, and foundation-model-based SLAM methods~\citep{orbslam2,ldso,droidslam,dpvslam,nicerslam,goslam,mast3rslam,vggtslam,vggtslam2,vggtslam++,SLAM-Former,slam3r,cut3r,zhuo2025streaming,lingbotmap}.
For reconstruction, we additionally include feed-forward reconstruction baselines~\citep{dust3r,mast3r,spann3r}.
Baseline results are taken from the corresponding papers or prior benchmark reports, as specified in the table captions.

All experiments are performed on a single NVIDIA A100 GPU (40GB VRAM) with an AMD EPYC 7542 CPU.
Runtime and memory usage are reported in Appendix~\ref{app:runtime}.
We use deterministic inference without random sampling or seed-dependent operations, and uncalibrated methods do not use ground-truth camera intrinsics.

Unless stated otherwise, \ours{} uses pretrained DROID-Net~\citep{droidslam} for high-frequency tracking.
For the low-frequency stream, per-image features are extracted with a pretrained DINO~\citep{oquab2023dinov2} backbone and aggregated across the window before Flow3r~\citep{flow3r} predicts pose, depth, and the displacement-confidence pairs described in~\cref{sec:factor_generation}.
Flow3r predictions initialize the SLAM state and, after state-consistency validation, refresh existing measurements in the dense BA backend described in~\cref{sec:method}.

\begin{table}[t]
\centering
\small
\caption{
Trajectory evaluation (ATE RMSE [m] $\downarrow$) on the
Replica dataset~\citep{replica}.
$*$ denotes results reproduced on our hardware.
}
\label{tab:replica_traj}

\resizebox{\textwidth}{!}{
\begin{tabular}{l|c|ccccccccc}
\toprule
Method & Uncalib.
& R0 & R1 & R2 & O0 & O1 & O2 & O3 & O4 & Avg. \\
\midrule

DROID-SLAM~\citep{droidslam}
& \xmark
& \sotabox{0.003}
& \sotabox{0.001}
& \sotabox{0.003}
& \secondbox{0.003}
& \secondbox{0.004}
& \sotabox{0.003}
& \secondbox{0.005}
& \sotabox{0.004}
& \sotabox{0.003} \\

NICER-SLAM~\citep{nicerslam}
& \xmark
& \secondbox{0.013}
& 0.016
& \secondbox{0.011}
& 0.021
& 0.032
& \secondbox{0.021}
& 0.014
& \secondbox{0.020}
& \secondbox{0.019} \\

\textbf{\ours{} (Ours)}
& \xmark
& \sotabox{0.003}
& \secondbox{0.002}
& \sotabox{0.003}
& \sotabox{0.002}
& \sotabox{0.003}
& \sotabox{0.003}
& \sotabox{0.003}
& \sotabox{0.004}
& \sotabox{0.003} \\

\midrule

VGGT-SLAM~\citep{vggtslam}
& \cmark
& \secondbox{0.030}
& 0.167
& 0.086
& 0.042
& 0.064
& 0.095
& 0.039
& 0.043
& 0.071 \\

VGGT-SLAM 2.0$^{*}$~\citep{vggtslam2}
& \cmark
& \secondbox{0.030}
& 0.049
& 0.040
& \secondbox{0.026}
& \secondbox{0.018}
& \secondbox{0.024}
& \secondbox{0.024}
& \secondbox{0.029}
& \secondbox{0.030} \\

SLAM-Former$^{*}$~\citep{SLAM-Former}
& \cmark
& 0.031
& \secondbox{0.033}
& \secondbox{0.025}
& 0.030
& 0.027
& 0.038
& 0.033
& 0.036
& 0.032 \\

SLAM3R~\citep{slam3r}
& \cmark
& 0.046
& 0.059
& 0.057
& 0.112
& 0.063
& 0.062
& 0.050
& 0.081
& 0.066 \\

CUT3R~\citep{cut3r}
& \cmark
& 0.145
& 0.243
& 0.127
& 0.159
& 0.230
& 0.162
& 0.088
& 0.204
& 0.170 \\

StreamVGGT~\citep{zhuo2025streaming}
& \cmark
& 0.113
& 0.163
& 0.077
& 0.076
& 0.070
& 0.180
& 0.153
& 0.168
& 0.125 \\

Lingbot-Map$^{*}$~\citep{lingbotmap}
& \cmark
& 0.202
& 0.165
& 0.104
& 0.156
& 0.115
& 0.166
& 0.101
& 0.213
& 0.153 \\

\textbf{\ours{} (Ours)}
& \cmark
& \sotabox{0.002}
& \sotabox{0.002}
& \sotabox{0.002}
& \sotabox{0.002}
& \sotabox{0.003}
& \sotabox{0.002}
& \sotabox{0.003}
& \sotabox{0.002}
& \sotabox{0.002} \\

\bottomrule
\end{tabular}
}
\end{table}

\begin{table*}[t]
\centering
\small
\caption{
Trajectory evaluation (ATE RMSE [m] $\downarrow$) on the
KITTI Odometry dataset~\citep{kitti}.
Baseline results are taken from VGGT-SLAM++~\citep{vggtslam++}.
}
\label{tab:kitti_traj}

\resizebox{\textwidth}{!}{
\begin{tabular}{l|c|ccccccccccc|c}
\toprule
Method & Uncalib.
& 00 & 01 & 02 & 03 & 04 & 05
& 06 & 07 & 08 & 09 & 10 & Avg. \\
\midrule

ORB-SLAM2~\citep{orbslam2}
& \xmark
& 40.65
& 502.20
& 47.82
& \sotabox{0.94}
& 1.30
& 29.95
& 40.82
& 16.04
& \sotabox{43.09}
& \secondbox{38.77}
& \sotabox{5.42}
& 69.73 \\

LDSO~\citep{ldso}
& \xmark
& \secondbox{9.32}
& \secondbox{11.68}
& \sotabox{31.98}
& 2.85
& 1.22
& \sotabox{5.10}
& \secondbox{13.55}
& \secondbox{2.96}
& 129.02
& \sotabox{21.64}
& 17.36
& \sotabox{22.43} \\

DROID-SLAM~\citep{droidslam}
& \xmark
& 92.10
& 5344.60
& 107.61
& \secondbox{2.38}
& 1.00
& 118.50
& 62.47
& 21.78
& 161.60
& 72.32
& 118.70
& 554.82 \\

DPV-SLAM~\citep{dpvslam}
& \xmark
& 112.80
& \sotabox{11.50}
& 123.53
& 2.50
& \secondbox{0.81}
& 57.80
& 54.86
& 18.77
& \secondbox{110.49}
& 76.66
& \secondbox{13.65}
& 53.03 \\

DPV-SLAM++~\citep{dpvslam}
& \xmark
& \sotabox{8.30}
& 11.86
& \secondbox{39.64}
& 2.50
& \sotabox{0.78}
& \secondbox{5.74}
& \sotabox{11.60}
& \sotabox{1.52}
& 110.90
& 76.70
& 13.70
& \secondbox{25.75} \\

\textbf{\ours{} (ours)}
& \xmark
& 117.08
& 64.77
& 134.50
& 3.94
& 1.69
& 83.39
& 65.43
& 25.37
& 112.10
& 102.82
& 27.95
& 67.19 \\

\midrule

VGGT-SLAM (Sim(3))~\citep{vggtslam}
& \cmark
& 125.11
& 120.96
& 288.82
& 5.16
& 0.96
& \secondbox{29.92}
& 15.03
& 14.58
& 235.80
& \secondbox{38.44}
& 18.60
& 81.22 \\


VGGT-SLAM++~\citep{vggtslam++}
& \cmark
& \secondbox{119.00}
& \secondbox{109.64}
& \secondbox{223.21}
& \secondbox{4.50}
& \secondbox{0.95}
& \sotabox{25.21}
& \sotabox{13.65}
& \secondbox{12.17}
& \secondbox{155.00}
& \sotabox{35.26}
& \secondbox{15.71}
& \secondbox{64.94} \\

\textbf{\ours{} (Ours)}
& \cmark
& \sotabox{39.99}
& \sotabox{58.15}
& \sotabox{28.09}
& \sotabox{1.78}
& \sotabox{0.73}
& 30.35
& \secondbox{14.49}
& \sotabox{6.87}
& \sotabox{75.98}
& 68.20
& \sotabox{11.36}
& \sotabox{30.18} \\

\bottomrule
\end{tabular}
}
\end{table*}
\begin{table*}[t]
\centering
\small
\setlength{\tabcolsep}{1.5pt}
\caption{
Reconstruction accuracy (Acc. [m] $\downarrow$) and completion
(Comp. [m] $\downarrow$) on the Replica dataset~\citep{replica}.
$+$ denotes results reported in NICER-SLAM,
$-$ denotes results reported in SLAM3R,
and $*$ denotes results reproduced on our hardware.
}
\label{tab:replica_recon}

\resizebox{\textwidth}{!}{%
\begin{tabular}{lcccccccccccccccccc}
\toprule
& \multicolumn{2}{c}{Room0}
& \multicolumn{2}{c}{Room1}
& \multicolumn{2}{c}{Room2}
& \multicolumn{2}{c}{Office0}
& \multicolumn{2}{c}{Office1}
& \multicolumn{2}{c}{Office2}
& \multicolumn{2}{c}{Office3}
& \multicolumn{2}{c}{Office4}
& \multicolumn{2}{c}{Avg.} \\

\cmidrule(lr){2-3}
\cmidrule(lr){4-5}
\cmidrule(lr){6-7}
\cmidrule(lr){8-9}
\cmidrule(lr){10-11}
\cmidrule(lr){12-13}
\cmidrule(lr){14-15}
\cmidrule(lr){16-17}
\cmidrule(lr){18-19}

Method
& Acc. & Comp.
& Acc. & Comp.
& Acc. & Comp.
& Acc. & Comp.
& Acc. & Comp.
& Acc. & Comp.
& Acc. & Comp.
& Acc. & Comp.
& Acc. & Comp. \\
\midrule

DROID-SLAM$^{+}$~\citep{droidslam}
& 0.1218 & 0.0896
& 0.0835 & 0.0607
& 0.0326 & 0.1601
& 0.0301 & 0.1619
& \secondbox{0.0239} & 0.1619
& 0.0566 & 0.1556
& 0.0449 & 0.0973
& 0.0465 & 0.0963
& 0.0550 & 0.1229 \\

NICER-SLAM$^{+}$~\citep{nicerslam}
& \sotabox{0.0253} & 0.0304
& 0.0393 & 0.0410
& 0.0340 & 0.0342
& 0.0549 & 0.0609
& 0.0345 & 0.0442
& 0.0402 & 0.0429
& 0.0334 & 0.0403
& \sotabox{0.0303} & 0.0387
& 0.0365 & 0.0416 \\

MASt3R-SLAM$^{*}$~\citep{mast3rslam}
& 0.2468 & 0.2675
& 0.0393 & 0.0266
& 0.0258 & 0.0187
& 0.0336 & 0.0211
& 0.0359 & 0.0303
& 0.0609 & 0.0484
& 0.0881 & 0.0824
& 0.0742 & 0.0577
& 0.0756 & 0.0691 \\

SLAM-Former$^{*}$~\citep{SLAM-Former}
& 0.0339 & 0.0247
& \sotabox{0.0248} & \secondbox{0.0171}
& \secondbox{0.0244} & \sotabox{0.0166}
& \sotabox{0.0208} & \secondbox{0.0153}
& \sotabox{0.0178} & \secondbox{0.0138}
& 0.0345 & 0.0222
& \sotabox{0.0296} & \secondbox{0.0213}
& \secondbox{0.0305} & \secondbox{0.0214}
& \secondbox{0.0270} & \secondbox{0.0187} \\

VGGT-SLAM 2.0$^{*}$~\citep{vggtslam2}
& \secondbox{0.0305} & 0.0622
& 0.0564 & 0.0765
& 0.0277 & 0.0629
& 0.0475 & 0.0457
& 0.0340 & 0.0595
& \sotabox{0.0249} & 0.0792
& \secondbox{0.0322} & 0.0739
& 0.0408 & 0.0743
& 0.0367 & 0.0657 \\

DUSt3R$^{-}$~\citep{dust3r}
& 0.0347 & 0.0250
& \secondbox{0.0253} & 0.0186
& 0.0295 & \secondbox{0.0176}
& 0.0492 & 0.0351
& 0.0309 & 0.0221
& 0.0401 & 0.0310
& 0.0327 & 0.0225
& 0.0366 & 0.0261
& 0.0349 & 0.0248 \\

MASt3R$^{-}$~\citep{mast3r}
& 0.0401 & 0.0410
& 0.0361 & 0.0325
& 0.0313 & 0.0215
& 0.0257 & 0.0163
& 0.1285 & 0.0813
& 0.0313 & \secondbox{0.0199}
& 0.0467 & 0.0315
& 0.0369 & 0.0247
& 0.0471 & 0.0336 \\

Spann3R$^{-}$~\citep{spann3r}
& 0.0975 & 0.1294
& 0.1551 & 0.1294
& 0.0728 & 0.0850
& 0.0546 & 0.1875
& 0.0524 & 0.1664
& 0.0933 & 0.1180
& 0.1600 & 0.0903
& 0.1397 & 0.1602
& 0.1032 & 0.1333 \\

SLAM3R$^{-}$~\citep{slam3r}
& 0.0319 & \secondbox{0.0240}
& 0.0312 & 0.0234
& 0.0272 & 0.0200
& 0.0428 & 0.0260
& 0.0317 & 0.0234
& 0.0384 & 0.0278
& 0.0390 & 0.0316
& 0.0432 & 0.0336
& 0.0357 & 0.0262 \\

\textbf{\ours{} (Ours)}
& 0.0313 & \sotabox{0.0226}
& 0.0259 & \sotabox{0.0166}
& \sotabox{0.0242} & \sotabox{0.0166}
& \secondbox{0.0230} & \sotabox{0.0148}
& \sotabox{0.0178} & \sotabox{0.0131}
& \secondbox{0.0258} & \sotabox{0.0169}
& 0.0327 & \sotabox{0.0210}
& 0.0321 & \sotabox{0.0210}
& \sotabox{0.0266} & \sotabox{0.0174} \\

\bottomrule
\end{tabular}%
}
\end{table*}















\begin{table}[t]
\begin{minipage}[t]{0.5\textwidth}
\centering
\small
\caption{
Reconstruction evaluation [m] on 7-Scenes dataset~\citep{shotton2013scene}.
Results are averaged over all sequences.
}
\label{tab:7scenes_recon}

\resizebox{0.92\linewidth}{!}{
\begin{tabular}{l|ccc}
\toprule
Method
& Acc. $\downarrow$
& Comp. $\downarrow$
& Chamfer $\downarrow$ \\
\midrule

MASt3R-SLAM~\citep{mast3rslam}
& 0.089 & 0.085 & 0.087 \\

VGGT-SLAM++~\citep{vggtslam++}
& 0.054 & 0.060 & 0.057 \\

SLAM-Former~\citep{SLAM-Former}
& \sotabox{0.024}
& 0.030
& \secondbox{0.027} \\

SLAM3R~\citep{slam3r}
& \secondbox{0.026}
& \secondbox{0.028}
& \secondbox{0.027} \\

\textbf{\ours{} (Ours)}
& \sotabox{0.024}
& \sotabox{0.015}
& \sotabox{0.019} \\

\bottomrule
\end{tabular}
}
\end{minipage}
\hfill
\begin{minipage}[t]{0.45\textwidth}
\centering
\small
\caption{
Ablation studies [m] on Replica dataset~\citep{replica}.
Results are averaged over all sequences.
}
\label{tab:ablation}

\resizebox{0.92\linewidth}{!}{
\begin{tabular}{l|l|ccc}
\toprule
 & Method
& ATE $\downarrow$
& Acc. $\downarrow$
& Comp. $\downarrow$ \\
\midrule

(a) & Basic Tracking
& 0.0027 & 0.0303 & 0.0188 \\

(b) & (a) + FF Init.
& 0.0026 & 0.0281 & 0.0177 \\

(c) & (a) + Meas. Refresh
& 0.0026 & 0.0291 & 0.0179 \\

(d) & Full w/o Gauge Align.
& 0.0024 & 0.0290 & 0.0194 \\

(e) & Full w/o Gate
& 0.5848 & 0.1889 & 0.2186 \\

\midrule

(f) & \textbf{Full (Ours)}
& \sotabox{0.0023}
& \sotabox{0.0266}
& \sotabox{0.0174} \\

\bottomrule
\end{tabular}
}
\end{minipage}
\end{table}

\subsection{Trajectory Evaluation}

As shown in~\cref{tab:tum_traj,tab:7scenes_traj,tab:replica_traj,tab:kitti_traj}, \ours{} achieves the best average result on three datasets and the second-best on 7-Scenes in the uncalibrated setting.

The strong result in Replica is consistent with favorable conditions for high-frequency DROID-Net tracking, including stable illumination, limited motion blur, and relatively small inter-frame motion.
The low-frequency stream supplies complementary multi-view geometry, consistent with the ablations in~\cref{tab:ablation}.

On KITTI Odometry, it substantially outperforms the evaluated uncalibrated foundation-model-based baselines, including VGGT-SLAM (Sim(3)) and VGGT-SLAM++, although some calibrated systems remain stronger.
In the calibrated setting, \ours{} improves substantially over plain DROID-SLAM ($554.82$\,m to $67.19$\,m average ATE), while it trails methods with explicit loop closure, such as LDSO and DPV-SLAM++, consistent with the absence of an explicit loop-closure mechanism in the current system (\cref{sec:conclusion}).



Additional qualitative trajectory visualizations are provided in Appendix~\ref{app:trajectory_visualization}.

\subsection{Reconstruction Evaluation}

\begin{figure}[t]
    \centering
    \includegraphics[width=0.95\linewidth]{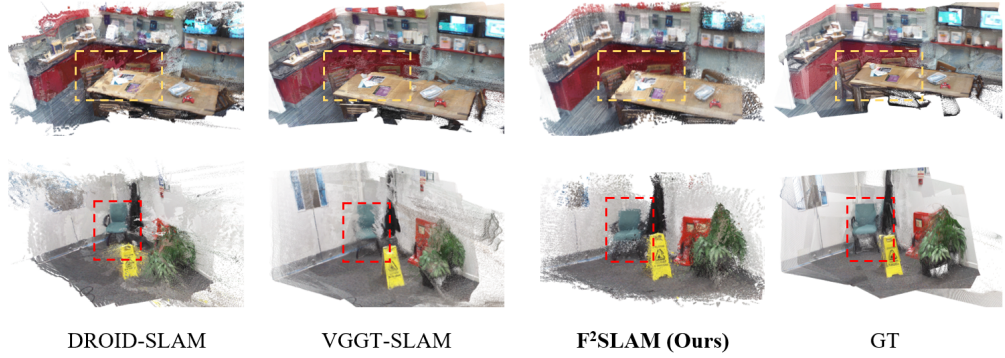}
    \caption{
    Qualitative reconstruction comparison on representative sequences.
    \ours{} recovers more continuous geometry with fewer missing regions and artifacts.
    }
    \label{fig:recon}
\end{figure}

We evaluate dense reconstruction on the Replica and 7-Scenes datasets, with quantitative results in~\cref{tab:7scenes_recon,tab:replica_recon} and qualitative comparisons in~\cref{fig:recon}.
For a fair comparison, we do not apply method-specific post-processing such as confidence-based point filtering~\citep{SLAM-Former}.

On Replica, \ours{} achieves the best average reconstruction accuracy and completion among the compared methods.
The lower completion error suggests that complementary low-frequency geometry helps recover regions weakly constrained by local tracking.

On 7-Scenes, \ours{} matches the best reconstruction accuracy at $0.024$\,m, while reducing completion error from $0.030$\,m to $0.015$\,m and Chamfer distance from $0.027$\,m to $0.019$\,m relative to SLAM-Former.
These correspond to approximately $50\%$ and $30\%$ improvements, respectively.
Qualitative results show the same trend, with more continuous surfaces and fewer missing regions.

Additional qualitative reconstruction results are provided in Appendix~\ref{app:qualitative_results}.

\subsection{Ablation}

\PAR{Intrinsic refinement.}
In the uncalibrated setting, \ours{} jointly optimizes camera intrinsics, poses, and inverse depths rather than fixing ground-truth calibration.
The paired results in~\cref{tab:tum_traj,tab:7scenes_traj,tab:replica_traj,tab:kitti_traj} show consistently lower average ATE, suggesting that joint refinement better adapts the geometric state to each sequence without requiring accurate calibration in advance.
These gains may also reflect compensation for systematic reprojection biases from resizing, cropping, rectification, or camera-model conventions, without implying inaccurate dataset calibration.

\PAR{Ablation of feed-forward integration.}
\cref{tab:ablation} evaluates feed-forward initialization, measurement refresh, gauge alignment, and state-consistent gating on Replica dataset.
Both initialization and measurement refresh independently improve performance over basic tracking.
The full model achieves the strongest overall performance, reducing ATE from $0.0027$\,m to $0.0023$\,m and improving reconstruction accuracy and completion from $0.0303$\,m and $0.0188$\,m to $0.0266$\,m and $0.0174$\,m, respectively.
Removing gauge alignment degrades both trajectory and reconstruction quality, highlighting the importance of expressing predictions in the current SLAM gauge.
Without state-consistent gating, performance degrades severely, demonstrating the need to reject inconsistent low-frequency predictions before measurement refresh. Further ablations on the low-frequency target-motion threshold, context window size, and gating threshold are provided in Appendix~\ref{sec:ablation_temporal_support}.
\section{Conclusion and Limitations}
\label{sec:conclusion}

\ours{} follows a single principle: feed-forward geometry should enter SLAM as optimization-native measurements rather than as a separate geometric state.
A high-frequency tracking stream and a low-frequency feed-forward stream convert their predictions into dense target-weight measurements on a persistent factor graph, allowing the same poses, inverse depths, and optional camera intrinsics to be jointly refined through a single dense BA backend with a state-consistency check.
The resulting system achieves strong trajectory estimation and dense reconstruction across the evaluated benchmarks.

\ours{} does not currently include explicit loop closure, which limits performance on long-horizon outdoor sequences such as KITTI Odometry. Performance also depends on feed-forward prediction reliability, which can degrade under weak texture, dynamic objects, large viewpoint changes, and scale ambiguity.
Incorporating loop closure and improving the robustness of low-frequency measurements are natural directions for future work.
\section*{AI use statement}
\label{sec:llm_usage}

Large language models (LLMs) were not used in the core methodology, experimental pipeline, scientific analysis, or claimed contributions of this work.
Specifically, LLMs were not involved in designing the proposed method, generating experimental results, selecting baselines, conducting evaluations, or drawing scientific conclusions.

LLM-based assistance was used only for language editing, such as improving grammar, clarity, and readability of the manuscript.
This assistance did not introduce new technical ideas, modify the method, change the experimental results, or affect the originality of the work.
All technical decisions, analyses, and claims were made by the authors, who take full responsibility for the content of the paper.
\bibliography{iclr2027_conference}
\bibliographystyle{iclr2027_conference}

\newpage
\appendix
\section{Appendix}

\subsection{Motion-based Keyframe and Window Scheduling}
\label{app:motion_keyframe}

\ours{} uses image-space motion to schedule keyframe insertion and low-frequency inference.
This avoids adding nearly duplicated frames to the persistent graph and adapts the temporal window to the observed camera motion.

Let $I_{k_n}$ denote the latest keyframe and $I_t$ the current frame.
The high-frequency stream estimates a dense displacement field $\mathbf{u}_{k_n t}$ from $I_{k_n}$ to $I_t$.
We define the average image-space motion as:
\begin{equation}
    m(k_n,t)
    =
    \frac{1}{|\Omega|}
    \sum_{\mathbf{p}\in\Omega}
    \left\|
        \mathbf{u}_{k_n t}(\mathbf{p})
    \right\|_2,
    \label{eq:f2slam_motion_score}
\end{equation}
where $\Omega$ denotes the valid pixel set.
The current frame is retained as a new keyframe when:
\begin{equation}
    m(k_n,t)
    >
    \tau_{\mathrm{kf}}.
    \label{eq:f2slam_keyframe_rule}
\end{equation}

Otherwise, it is used only for transient tracking and is not added as a persistent graph node.

Newly inserted keyframes are accumulated in a low-frequency buffer:
\begin{equation}
    \mathcal{B}_t
    =
    \{k_s,\ldots,k_e\}.
    \label{eq:f2slam_low_buffer}
\end{equation}

The accumulated image-space motion of the buffer is:
\begin{equation}
    M(\mathcal{B}_t)
    =
    \sum_{i=s}^{e-1}
    m(k_i,k_{i+1}).
    \label{eq:f2slam_acc_motion}
\end{equation}

This quantity is used only as a scheduling signal and does not represent metric camera displacement.

Low-frequency inference is triggered when the buffer contains sufficient temporal context and its accumulated motion exceeds a predefined threshold.
A maximum buffer length is additionally imposed to bound latency in slow-motion sequences. The corresponding thresholds are specified in the implementation details.

Once triggered, the current buffer forms the low-frequency window:
\begin{equation}
    \mathcal{W}_t
    =
    \mathcal{B}_t.
    \label{eq:f2slam_scheduled_window}
\end{equation}
The window does not define an independent submap or an additional optimized state.
Instead, it determines which existing active edges are eligible for low-frequency factor generation and subsequent gating:
\begin{equation}
    \mathcal{P}_{\mathrm{low}}^t
    =
    \left\{
        (i,j)\in\mathcal{E}_{\mathrm{act}}^t
        \;\middle|\;
        i,j\in\mathcal{W}_t
    \right\}.
    \label{eq:f2slam_scheduled_pairs}
\end{equation}

Accepted measurements refresh the corresponding target-weight factor-states, as described in~\cref{sec:graph_update}.
Thus, motion determines when the low-frequency stream is invoked, while both streams share the same graph topology and optimized geometric state.

\subsection{State Optimization with Target-Weight Factor-States}
\label{app:state_optimization}

\ours{} maintains and optimizes a single persistent geometric state using confidence-weighted dense reprojection constraints.
For the current keyframe set $\mathcal{K}_t$, the optimized state is:
\begin{equation}
    \mathcal{X}_t
    =
    \left\{
        \mathbf{T}_i,
        \mathbf{d}_i
    \right\}_{i\in\mathcal{K}_t}
    \cup
    \left\{
        \boldsymbol{\theta}
    \right\},
    \label{eq:app_state}
\end{equation}
where $\mathbf{T}_i\in SE(3)$ denotes the camera pose of keyframe $i$, and $\mathbf{d}_i\in\mathbb{R}_{+}^{h\times w}$ denotes its dense inverse-depth map.
The camera intrinsics are parameterized as:
\begin{equation}
    \boldsymbol{\theta}
    =
    \left(
        f_x,
        f_y,
        c_x,
        c_y
    \right)^{\top}.
    \label{eq:app_intrinsics}
\end{equation}

When the camera calibration is known, $\boldsymbol{\theta}$ is fixed.
Otherwise, it is refined jointly with the camera poses and inverse-depth maps.

Each active directed edge $(i,j)\in\mathcal{E}_{\mathrm{act}}^t$
stores a mutable factor-state:
\begin{equation}
    \mathcal{F}_{ij}
    =
    \left(
        \mathcal{M}_{ij},
        \ell_{ij},
        \gamma_{ij}
    \right),
    \label{eq:app_factor_state}
\end{equation}
where
\begin{equation}
    \mathcal{M}_{ij}
    =
    \left(
        \mathbf{p}^{*}_{ij},
        \mathbf{w}_{ij}
    \right)
    \label{eq:app_target_weight_measurement}
\end{equation}
is the current target-weight measurement.
Here, $\mathbf{p}^{*}_{ij}$ is the dense target-coordinate field, $\mathbf{w}_{ij}$ is its per-pixel precision weight, $\ell_{ij}\in\{\mathrm{high},\mathrm{low}\}$ records the latest measurement source, and $\gamma_{ij}\in\{0,1\}$ is a temporary update lock.

During each bundle-adjustment step, $\mathcal{M}_{ij}$ is treated as a fixed observation.
The source label $\ell_{ij}$ and lock flag $\gamma_{ij}$ control factor-state updates but do not enter the optimization objective.

For a source pixel $\mathbf{p}\in\Omega_i$, the current state predicts its target coordinate in keyframe $j$ as:
\begin{equation}
    \hat{\mathbf{p}}_{ij}
    \left(
        \mathbf{p};
        \mathcal{X}_t
    \right)
    =
    \pi_{\boldsymbol{\theta}}
    \left(
        \mathbf{T}_{j}
        \mathbf{T}_{i}^{-1}
        \,
        \pi_{\boldsymbol{\theta}}^{-1}
        \left(
            \mathbf{p},
            \mathbf{d}_i(\mathbf{p})
        \right)
    \right),
    \label{eq:app_predicted_target}
\end{equation}
where $\pi_{\boldsymbol{\theta}}$ and $\pi_{\boldsymbol{\theta}}^{-1}$ denote projection and inverse projection under the camera model specified by $\boldsymbol{\theta}$.

The corresponding dense target-coordinate residual is:
\begin{equation}
    \mathbf{r}_{ij}
    \left(
        \mathbf{p};
        \mathcal{X}_t
    \right)
    =
    \hat{\mathbf{p}}_{ij}
    \left(
        \mathbf{p};
        \mathcal{X}_t
    \right)
    -
    \mathbf{p}^{*}_{ij}(\mathbf{p}).
    \label{eq:app_target_residual}
\end{equation}

The per-pixel precision defines the diagonal information matrix:
\begin{equation}
    \mathbf{\Lambda}_{ij}(\mathbf{p})
    =
    \operatorname{diag}
    \left(
        \mathbf{w}_{ij}(\mathbf{p})
    \right).
    \label{eq:app_weight_matrix}
\end{equation}

The persistent state is optimized by minimizing:
\begin{equation}
\begin{aligned}
    \mathcal{X}_t^{*}
    =
    \arg\min_{\mathcal{X}_t}
    \;&
    \sum_{(i,j)\in\mathcal{E}_{\mathrm{BA}}^t}
    \sum_{\mathbf{p}\in\Omega_i}
    v_{ij}(\mathbf{p})
    \,
    \rho
    \left(
        \mathbf{r}_{ij}
        \left(
            \mathbf{p};
            \mathcal{X}_t
        \right)^{\top}
        \mathbf{\Lambda}_{ij}(\mathbf{p})
        \mathbf{r}_{ij}
        \left(
            \mathbf{p};
            \mathcal{X}_t
        \right)
    \right)
    \\
    &+
    \frac{\lambda_d}{2}
    \sum_{i\in\mathcal{K}_t}
    \left\|
        \nabla\mathbf{d}_i
    \right\|_2^2
    \\
    &+
    \frac{\lambda_{\theta}}{2}
    \left\|
        \boldsymbol{\theta}
        -
        \boldsymbol{\theta}_0
    \right\|_{\mathbf{\Sigma}_{\theta}^{-1}}^2 .
\end{aligned}
    \label{eq:app_state_objective}
\end{equation}
where $v_{ij}(\mathbf{p})\in\{0,1\}$ masks invalid projections, $\rho(\cdot)$ is a robust penalty, and $\lambda_d$ and $\lambda_{\theta}$ control the strengths of the inverse-depth and intrinsic regularization terms, respectively.
The intrinsic prior is omitted when the camera calibration is fixed.

The first term in Eq.~\eqref{eq:app_state_objective} enforces agreement between the current geometric state and the target-coordinate measurements stored on all active edges.
The second term regularizes spatial variation in the dense inverse-depth maps. The third term constrains the optimized intrinsics around the calibration prior $\boldsymbol{\theta}_0$, with uncertainty described by $\mathbf{\Sigma}_{\theta}$.

For compact notation, we stack all inverse-depth maps as:
\begin{equation}
    \mathbf{d}
    =
    \left[
        \mathbf{d}_1^{\top},
        \ldots,
        \mathbf{d}_{N_t}^{\top}
    \right]^{\top},
    \label{eq:app_stacked_depth}
\end{equation}
and denote the corresponding block-diagonal spatial-gradient operator by $\mathbf{D}$.
The inverse-depth regularizer can then be written as:
\begin{equation}
    \mathcal{R}_{d}(\mathbf{d})
    =
    \frac{\lambda_d}{2}
    \left\|
        \mathbf{D}\mathbf{d}
    \right\|_2^2.
    \label{eq:app_depth_regularizer}
\end{equation}

Similarly, the intrinsic prior is:
\begin{equation}
    \mathcal{R}_{\theta}(\boldsymbol{\theta})
    =
    \frac{\lambda_{\theta}}{2}
    \left\|
        \boldsymbol{\theta}
        -
        \boldsymbol{\theta}_0
    \right\|_{\mathbf{\Sigma}_{\theta}^{-1}}^2.
    \label{eq:app_intrinsic_regularizer}
\end{equation}

At the current estimate, the residual for each correspondence is linearized as:
\begin{equation}
\begin{aligned}
    \mathbf{r}_{ij}
    \left(
        \mathbf{p};
        \mathcal{X}_t
        \boxplus
        \Delta
    \right)
    \approx\;&
    \mathbf{r}_{ij}
    \left(
        \mathbf{p};
        \mathcal{X}_t
    \right)
    +
    \mathbf{J}^{\xi_i}_{ij}(\mathbf{p})
    \delta\boldsymbol{\xi}_i
    \\
    &+
    \mathbf{J}^{\xi_j}_{ij}(\mathbf{p})
    \delta\boldsymbol{\xi}_j
    +
    \mathbf{J}^{d}_{ij}(\mathbf{p})
    \delta\mathbf{d}_i(\mathbf{p})
    \\
    &+
    \mathbf{J}^{\theta}_{ij}(\mathbf{p})
    \delta\boldsymbol{\theta},
\end{aligned}
    \label{eq:app_state_linearization}
\end{equation}
where $\delta\boldsymbol{\xi}_i\in\mathbb{R}^{6}$ is the pose increment in the Lie algebra of $SE(3)$, $\delta\mathbf{d}_i$ is the inverse-depth increment, and $\delta\boldsymbol{\theta}$ is the intrinsic increment.
When the intrinsics are fixed, the final term is omitted.

Using iteratively reweighted least squares for the robust penalty, all active residuals are stacked into:
\begin{equation}
    \mathbf{r}
    +
    \mathbf{J}\Delta,
    \label{eq:app_stacked_linearization}
\end{equation}
where $\widetilde{\mathbf{W}}$ combines the target precision, projection-validity masks, and robust weights.

The regularization terms contribute:
\begin{equation}
    \mathbf{H}_{d}
    =
    \lambda_d
    \mathbf{D}^{\top}
    \mathbf{D},
    \qquad
    \mathbf{g}_{d}
    =
    \lambda_d
    \mathbf{D}^{\top}
    \mathbf{D}
    \mathbf{d},
    \label{eq:app_depth_normal_terms}
\end{equation}
and
\begin{equation}
    \mathbf{H}_{\theta}
    =
    \lambda_{\theta}
    \mathbf{\Sigma}_{\theta}^{-1},
    \qquad
    \mathbf{g}_{\theta}
    =
    \lambda_{\theta}
    \mathbf{\Sigma}_{\theta}^{-1}
    \left(
        \boldsymbol{\theta}
        -
        \boldsymbol{\theta}_0
    \right).
    \label{eq:app_intrinsic_normal_terms}
\end{equation}

These terms are inserted into the blocks corresponding to the inverse-depth and intrinsic variables, respectively.

The resulting damped Gauss--Newton system is:
\begin{equation}
\begin{aligned}
    \bigl(
        \mathbf{J}^{\top}
        \widetilde{\mathbf{W}}
        \mathbf{J}
        +
        \mathbf{H}_{d}
        +
        \mathbf{H}_{\theta}
        +
        \lambda_{\mathrm{lm}}
        \mathbf{I}
    \bigr)
    \Delta
    =
    -
    \bigl(
        \mathbf{J}^{\top}
        \widetilde{\mathbf{W}}
        \mathbf{r}
        +
        \mathbf{g}_{d}
        +
        \mathbf{g}_{\theta}
    \bigr),
\end{aligned}
    \label{eq:app_gn_system}
\end{equation}
where $\lambda_{\mathrm{lm}}$ is the damping coefficient.

The complete increment vector is:
\begin{equation}
    \Delta
    =
    \left[
        \delta\boldsymbol{\xi}_1^{\top},
        \ldots,
        \delta\boldsymbol{\xi}_{N_t}^{\top},
        \delta\mathbf{d}_1^{\top},
        \ldots,
        \delta\mathbf{d}_{N_t}^{\top},
        \delta\boldsymbol{\theta}^{\top}
    \right]^{\top}.
    \label{eq:app_increment}
\end{equation}

When the camera intrinsics are fixed, $\delta\boldsymbol{\theta}$ and the corresponding Jacobian, gradient, and Hessian blocks are removed.

After solving Eq.~\eqref{eq:app_gn_system}, the pose variables are updated using a right-multiplicative Lie-group update:
\begin{equation}
    \mathbf{T}_{i}
    \leftarrow
    \mathbf{T}_{i}
    \exp
    \left(
        \delta\boldsymbol{\xi}_{i}^{\wedge}
    \right).
    \label{eq:app_pose_update}
\end{equation}

The inverse-depth maps are updated additively:
\begin{equation}
    \mathbf{d}_{i}
    \leftarrow
    \mathbf{d}_{i}
    +
    \delta\mathbf{d}_{i}.
    \label{eq:app_depth_update}
\end{equation}

When jointly optimized, the camera intrinsics are updated as:
\begin{equation}
    \boldsymbol{\theta}
    \leftarrow
    \boldsymbol{\theta}
    +
    \delta\boldsymbol{\theta}.
    \label{eq:app_intrinsic_update}
\end{equation}

Low-frequency information enters this optimization exclusively through the target-weight measurements stored on selected active edges.
When a candidate passes the state-consistency gating procedure described in~\cref{sec:graph_update}, it refreshes the current measurement $\mathcal{M}_{ij}$ on the corresponding edge. Rejected candidates leave the existing measurement unchanged, while all other active edges retain their current high-frequency measurements.

Importantly, refreshing an edge changes neither the graph topology nor the optimized variable set.
Each active edge contributes exactly one dense target-weight residual, regardless of whether its current measurement was generated by the high-frequency stream or accepted from the low-frequency stream.
The same bundle-adjustment objective and solver therefore optimize one persistent SLAM state using a unified set of high-frequency and accepted low-frequency constraints.





\subsection{Experimental Details}

We evaluate all methods at a common set of ground-truth timestamps for consistent comparison.
Estimated poses are associated with the corresponding ground-truth poses by timestamp, and all trajectories are aligned to the ground truth using a similarity transformation in $\mathrm{Sim}(3)$.
All reported trajectory errors are then computed from the aligned trajectories under the same evaluation protocol.

All methods are evaluated on the same input sequences and ground-truth trajectories.
Unless otherwise specified, we use the original image resolution and the default configuration of each baseline, while preserving its original keyframe-selection and frame-processing strategy.

For \ours{}, we use the same configuration across all datasets. 
Keyframes are inserted with an image-space motion threshold of $\tau_{\mathrm{kf}}=40$.
The low-frequency stream is triggered at a target motion of $800$ and uses a maximum context size of $20$ keyframes.
For the window-size ablation, this maximum is varied from $4$ to $40$ while all other parameters remain fixed.
For state-consistent admission, we set $\tau_{\mathrm{sup}}=0.05$, $\tau_{\mathrm{geo}}=3.0$, and $\eta_{\mathrm{low}}=0.25$.
Accepted low-frequency measurements remain locked while active and are removed after $20$ frontend updates.
The high-frequency graph uses $3$ temporal and $2$ proximity neighbors, and dense BA performs $2$ iterations per graph update.

\subsection{Runtime Analysis}
\label{app:runtime}

For SLAM systems, computational efficiency is also an important consideration.
We therefore compare the runtime performance of different methods on the Replica dataset.
All methods are evaluated on the same NVIDIA A100 GPU with 40\,GB of memory under the same hardware environment.

We report the average processing frame rate in frames per second (FPS) together with peak GPU memory usage.
FPS is computed based on the number of processed input frames and the corresponding processing time, and the reported results are averaged over the evaluated Replica sequences.

\begin{table}[t]
\begin{minipage}[t]{0.63\textwidth}
\centering
\small
\caption{Runtime performance comparison on the Replica dataset. All methods are evaluated on the same NVIDIA A100 (40\,GB) GPU.}
\label{tab:runtime}
\resizebox{\linewidth}{!}{
\begin{tabular}{lcc}
\toprule
Method              & FPS $\uparrow$   & Peak GPU [MB] $\downarrow$ \\
\midrule
SLAM-Former   & \sotabox{8.154} & 27693 \\
VGGT-SLAM 2.0   & 3.909 & \sotabox{6523}  \\
\textbf{\ours{} (Ours)}           & \secondbox{7.399} & \secondbox{15443} \\
\bottomrule
\end{tabular}
}
\end{minipage}
\hfill
\begin{minipage}[t]{0.31\textwidth}
\centering
\small
\caption{Sensitivity to the geometric consistency threshold on Replica.}
\label{tab:gate_sensitivity}
\resizebox{\linewidth}{!}{
\begin{tabular}{cc}
\toprule
$\tau_{\mathrm{geo}}$ & ATE RMSE $\downarrow$ \\
\midrule
2.0 & 0.25 \\
3.0 & \textbf{0.23} \\
4.0 & 0.24 \\
\bottomrule
\end{tabular}
}
\end{minipage}
\end{table}

As shown in~\cref{tab:runtime}, \ours{} achieves competitive computational efficiency, reaching $7.399$ FPS with moderate GPU memory usage.
Compared with SLAM-Former, our method achieves a similar processing speed while reducing peak GPU memory consumption.
It also provides substantially higher processing throughput than VGGT-SLAM 2.0.
These results demonstrate that \ours{} maintains an effective balance between localization performance and computational efficiency.

\subsection{Ablation on Low-frequency Temporal Support and Gating}
\label{sec:ablation_temporal_support}

We study two parameters that control the temporal support of the low-frequency stream: the target motion used to trigger inference and the number of keyframes included in each window.
The target-motion threshold determines when sufficient camera displacement has accumulated for a low-frequency update, whereas the window size controls how much temporal context is jointly processed.

\begin{table}[t]
\centering
\small
\caption{Ablation study of Flow3r target motion on 7-Scenes. We report Sim(3)-aligned ATE RMSE (m). Lower is better.}
\label{tab:7scenes_motion_ablation}
\resizebox{\textwidth}{!}{
\begin{tabular}{l|c|ccccccc|c}
\toprule
Method & Opt. Intr. & Chess & Fire & Heads & Office & Pumpkin & RedKitchen & Stairs & Avg. \\
\midrule
w/ target motion 600
& \cmark
& 0.036 & 0.024 & 0.014 & 0.092 & 0.143 & 0.048 & 0.025 & 0.054 \\

w/ target motion 800 (Ours)
& \cmark
& \textbf{0.036} & \textbf{0.024} & \textbf{0.014} & \textbf{0.087} & \textbf{0.140} & \textbf{0.041} & \textbf{0.020} & \textbf{0.051} \\

w/ target motion 1000
& \cmark
& 0.036 & 0.025 & 0.014 & 0.087 & 0.140 & 0.041 & 0.020 & 0.051 \\
\bottomrule
\end{tabular}
}
\end{table}

\begin{figure}[t]
    \centering
    \includegraphics[width=0.80\linewidth]{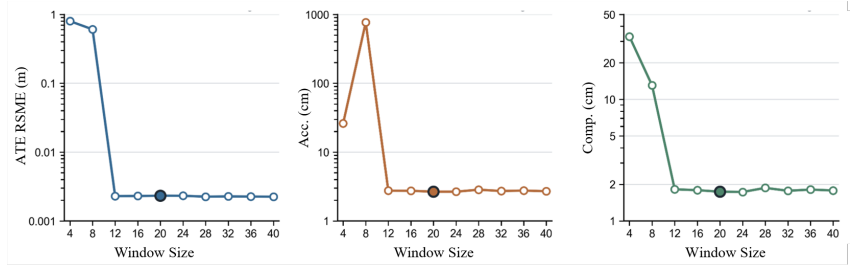}
    \caption{
    Ablation on the low-frequency context window size.
    Lower is better for all metrics.
    Log-scaled y-axes retain a continuous line plot while keeping the large errors at sizes 4 and 8 visible.
    We use a logarithmic coordinate system to better show the overall trend.
    }
    \label{fig:window_ablation}
\end{figure}

\PAR{Target motion.}
\cref{tab:7scenes_motion_ablation} shows that increasing the target motion from $600$ to $800$ reduces the average ATE RMSE from $0.054$\,m to
$0.051$\,m.
The improvement is most evident on Office, RedKitchen, and Stairs, where a larger motion span provides additional viewpoint diversity beyond the short-range constraints of the high-frequency stream.
This suggests that the low-frequency update is most useful when it captures sufficiently separated observations rather than repeatedly processing nearly redundant local views.

Increasing the threshold further to $1000$ does not provide a consistent additional gain, with the average ATE remaining at $0.051$\,m.
At such larger motion spans, the stronger geometric baseline is likely balanced by more severe viewpoint, occlusion, and appearance changes, which can reduce the reliability of the predicted measurements.
Importantly, performance does not noticeably degrade, indicating that the candidate gating mechanism prevents unreliable low-frequency measurements from strongly perturbing the active graph.
We therefore adopt 800 as the default target motion.

\PAR{Window size.}
\cref{fig:window_ablation} evaluates maximum context sizes from $4$ to $40$ keyframes while keeping the data, optimizer, trigger threshold, and gating rules fixed. The two smallest settings produce catastrophic failures because their temporal support is insufficient to establish reliable multi-view geometry. 
With only a few keyframes, the window provides limited baseline and viewpoint diversity, making correspondence and depth predictions poorly constrained. 
The resulting low-frequency measurements are therefore noisy and can be rejected inconsistently by the state-consistency gate. 
This effect is especially severe in the smallest windows, where a small change in the sampled views can substantially alter the available geometric evidence.

Once the window reaches a moderate size ($12$ keyframes and above), it contains enough overlapping yet non-redundant views to support stable correspondence, depth, and confidence estimation.
Candidate gating then suppresses residual outliers before they refresh graph factors, which explains the sharp recovery and the relatively flat, non-monotonic curves in the remaining range.
Adding more keyframes beyond this regime mostly introduces redundant observations and does not yield a consistent geometric gain.
We therefore use $20$ keyframes as the default context size: it attains the best Accuracy while remaining close to the best completion and within the stable ATE regime.
This is a balanced choice across the three metrics rather than a claim of uniformly optimal performance.

Overall, these ablations show that \ours{} requires a minimum amount of low-frequency temporal support, after which performance is relatively insensitive to the exact window size. 
This robustness is enabled by candidate gating and measurement-level factor refresh, which allow measurements conditioned on broader multi-view context to update the graph while suppressing unreliable updates.

\PAR{Gating threshold.}
\cref{tab:gate_sensitivity} evaluates the geometric consistency threshold $\tau_{\mathrm{geo}}$ around its default value of $3.0$.
Performance remains stable across the tested range.
A stricter threshold may reject useful feed-forward measurements, whereas a more permissive threshold may admit less reliable geometric evidence.
The default setting provides a balanced operating point between measurement coverage and geometric consistency.

Overall, these ablations show that \ours{} remains robust to moderate variations in its low-frequency configuration.
The target-motion and context-window experiments demonstrate stable behavior across different temporal support, while the gating analysis indicates that state-consistent admission does not rely on a narrowly tuned threshold.

\subsection{Trajectory Visualization}
\label{app:trajectory_visualization}

We provide qualitative trajectory comparisons to complement the quantitative ATE evaluation.
The visualizations examine trajectory accuracy from three perspectives: overall sequence-level behavior, two-dimensional projected motion, and full three-dimensional geometry.

\begin{figure*}[t]
    \centering
    \includegraphics[width=0.98\linewidth]{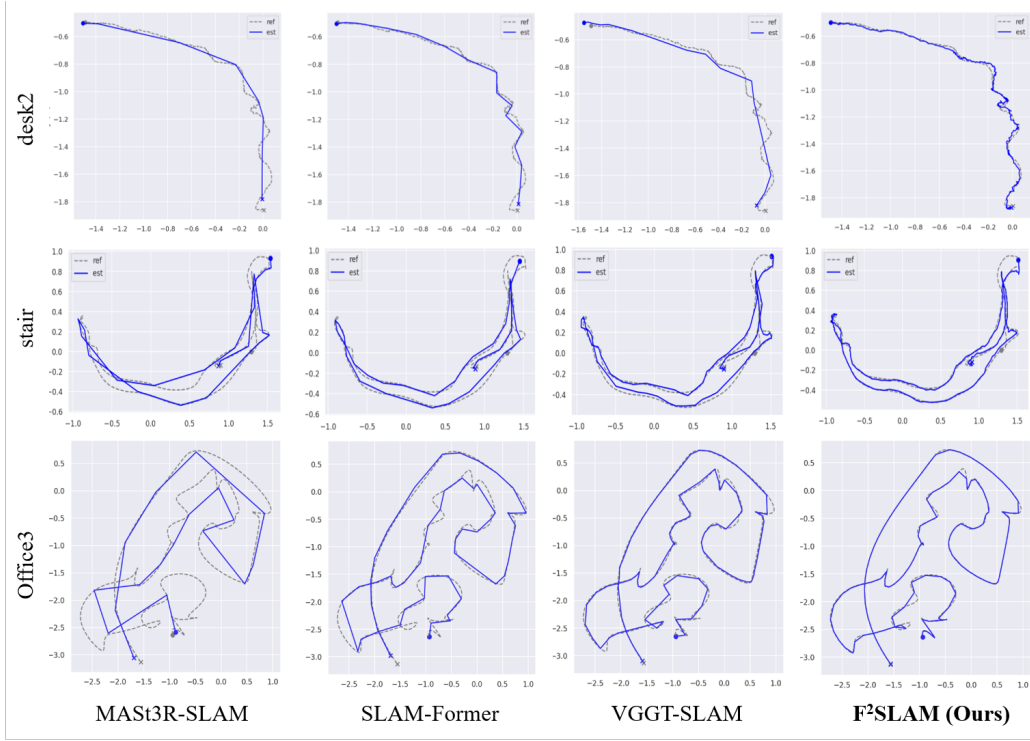}
    \caption{
    Qualitative trajectory comparison on representative sequences.
    All estimated trajectories are aligned to the ground truth using the same evaluation protocol.
    Compared with competing methods, \ours{} follows the reference
    trajectory more closely throughout the sequence and exhibits less accumulated drift.
    }
    \label{fig:trajectory_compare}
\end{figure*}

\cref{fig:trajectory_compare} provides an overall comparison across representative sequences.
Several competing methods recover short-term motion reasonably well, but small local errors accumulate into visible global deformation, endpoint offset, or trajectory-scale inconsistency.
In contrast, \ours{} remains close to the ground truth over a larger portion of each sequence and preserves the global trajectory structure more faithfully.
This shows that the improvement is not limited to isolated poses or final endpoint alignment, but extends across the complete keyframe trajectory.

\begin{figure*}[t]
    \centering
    \includegraphics[width=0.98\linewidth]{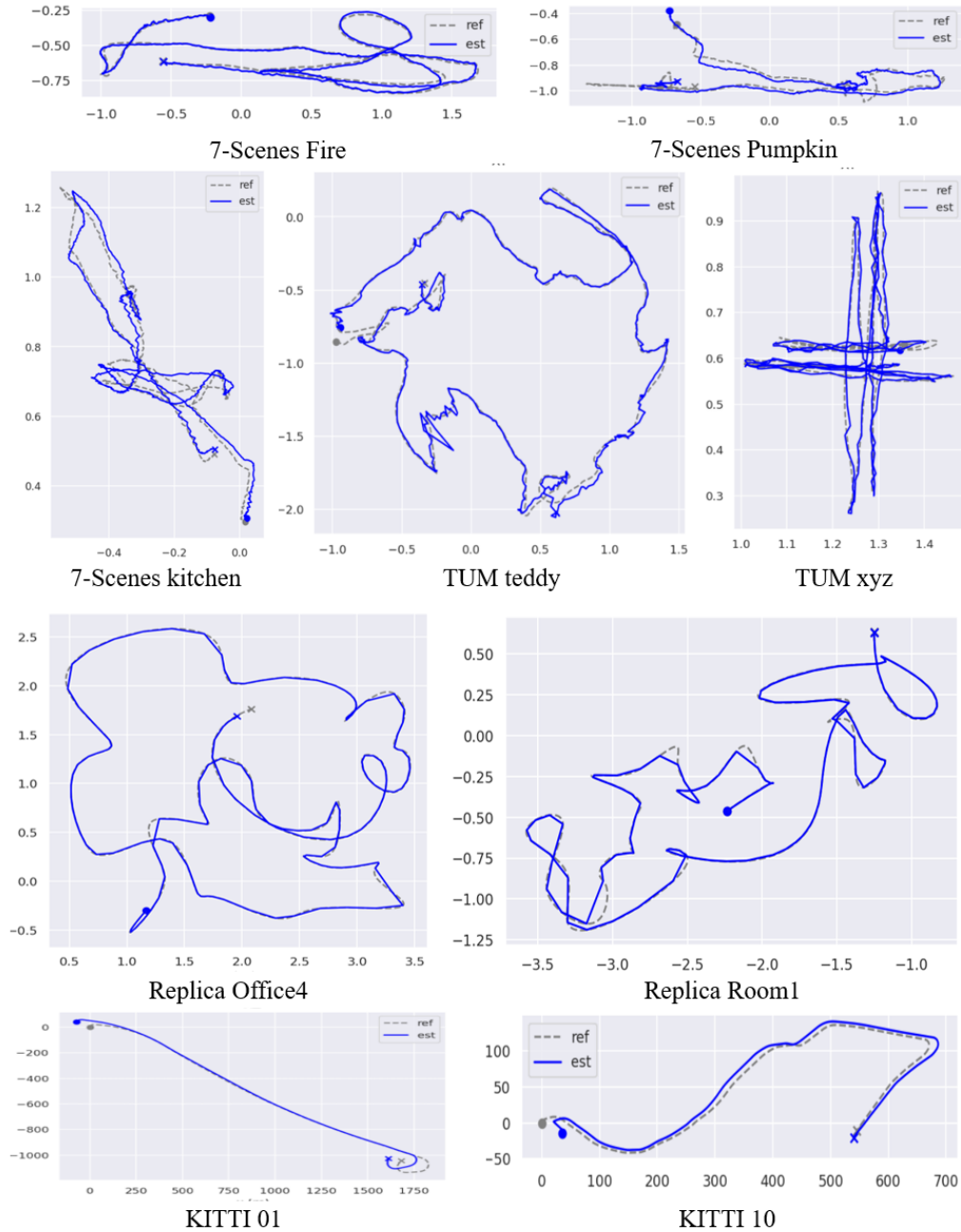}
    \caption{
    Two-dimensional trajectory projections on representative sequences.
    The projections emphasize global shape, turning behavior, and endpoint drift.
    \ours{} better preserves the reference path and avoids the progressive deviation observed in competing methods.
    }
    \label{fig:trajectory_2d}
\end{figure*}

\cref{fig:trajectory_2d} projects the trajectories onto a representative two-dimensional plane, making accumulated translation and orientation errors easier to identify.
Although the baselines often reproduce the coarse motion pattern, their estimates progressively deviate around turns and long trajectory segments.
\ours{} more accurately preserves both the local turning structure and the global path shape, resulting in smaller endpoint displacement and less systematic drift.

\begin{figure*}[t]
    \centering
    \includegraphics[width=0.98\linewidth]{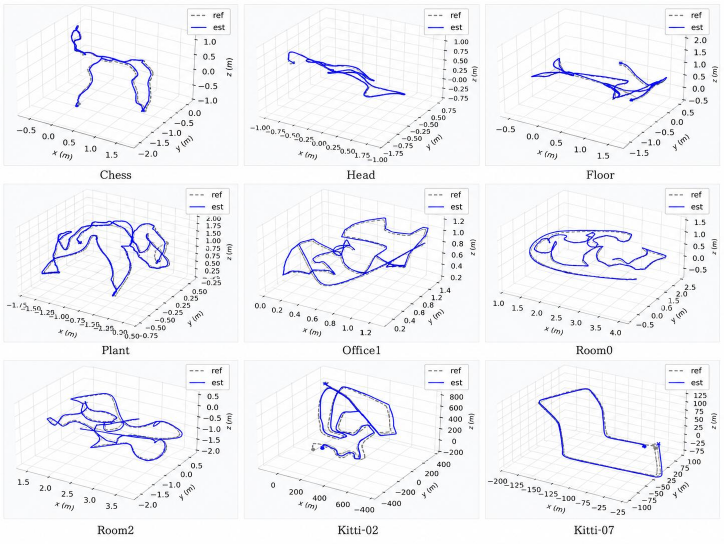}
    \caption{
    Three-dimensional trajectory visualization.
    The full spatial trajectories reveal vertical drift and geometric distortion that can be hidden by planar projections.
    \ours{} maintains a more coherent trajectory across all spatial directions.
    }
    \label{fig:trajectory_3d}
\end{figure*}

\cref{fig:trajectory_3d} further compares the trajectories in full three-dimensional space.
This view exposes vertical drift, out-of-plane deviations, and geometric distortions that may appear small in a two-dimensional projection.
The baseline trajectories exhibit larger spatial inconsistency as the sequence progresses, whereas \ours{} remains more coherent in all three dimensions and better preserves the overall motion geometry.

These qualitative results support the quantitative findings and clarify the source of the improvement.
High-frequency factors maintain accurate local tracking, while accepted low-frequency measurements predicted from broader multi-view context repeatedly correct accumulated global drift within the persistent graph.
As a result, \ours{} achieves not only lower endpoint error, but also better trajectory shape, stronger long-term consistency, and more stable motion estimates throughout the sequence.

\subsection{Additional Qualitative Results}
\label{app:qualitative_results}

\begin{figure*}[t]
    \centering
    \includegraphics[width=0.92\textwidth]{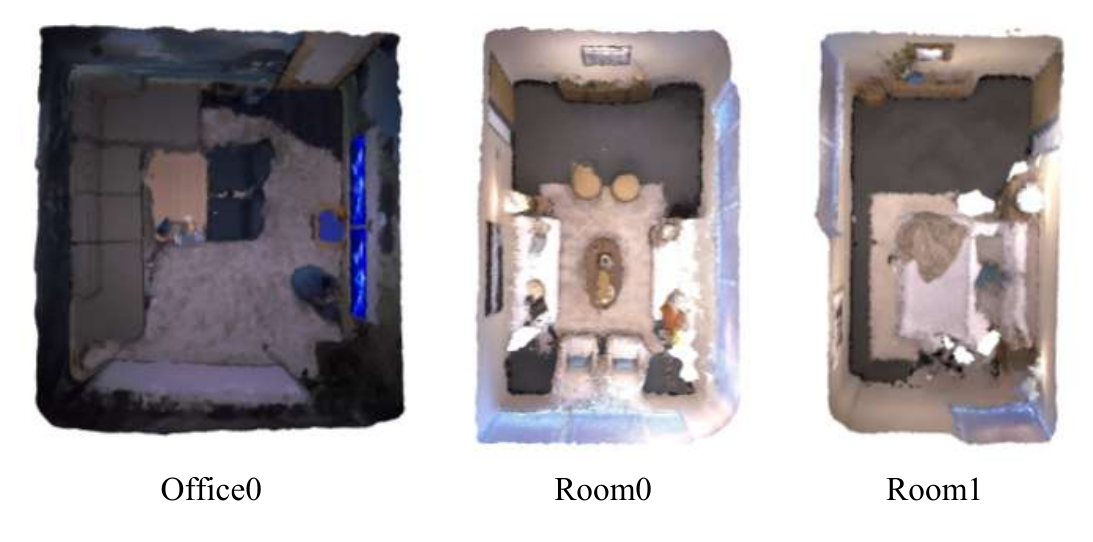}
    \caption{
        Qualitative reconstruction results on representative Replica scenes.
        \ours{} recovers coherent room layouts and preserves major scene structures across long sequences.
    }
    \label{fig:replica_appendix}
\end{figure*}

We provide additional qualitative results on Replica and KITTI to complement the quantitative evaluation in the main paper.
\cref{fig:replica_appendix} shows reconstructions of representative Replica scenes, including \textit{Office0}, \textit{Room0}, and \textit{Room1}.
\ours{} recovers coherent room layouts and preserves the spatial arrangement of major scene structures across the sequences.

\begin{figure*}[t]
    \centering
    \includegraphics[width=0.96\textwidth]{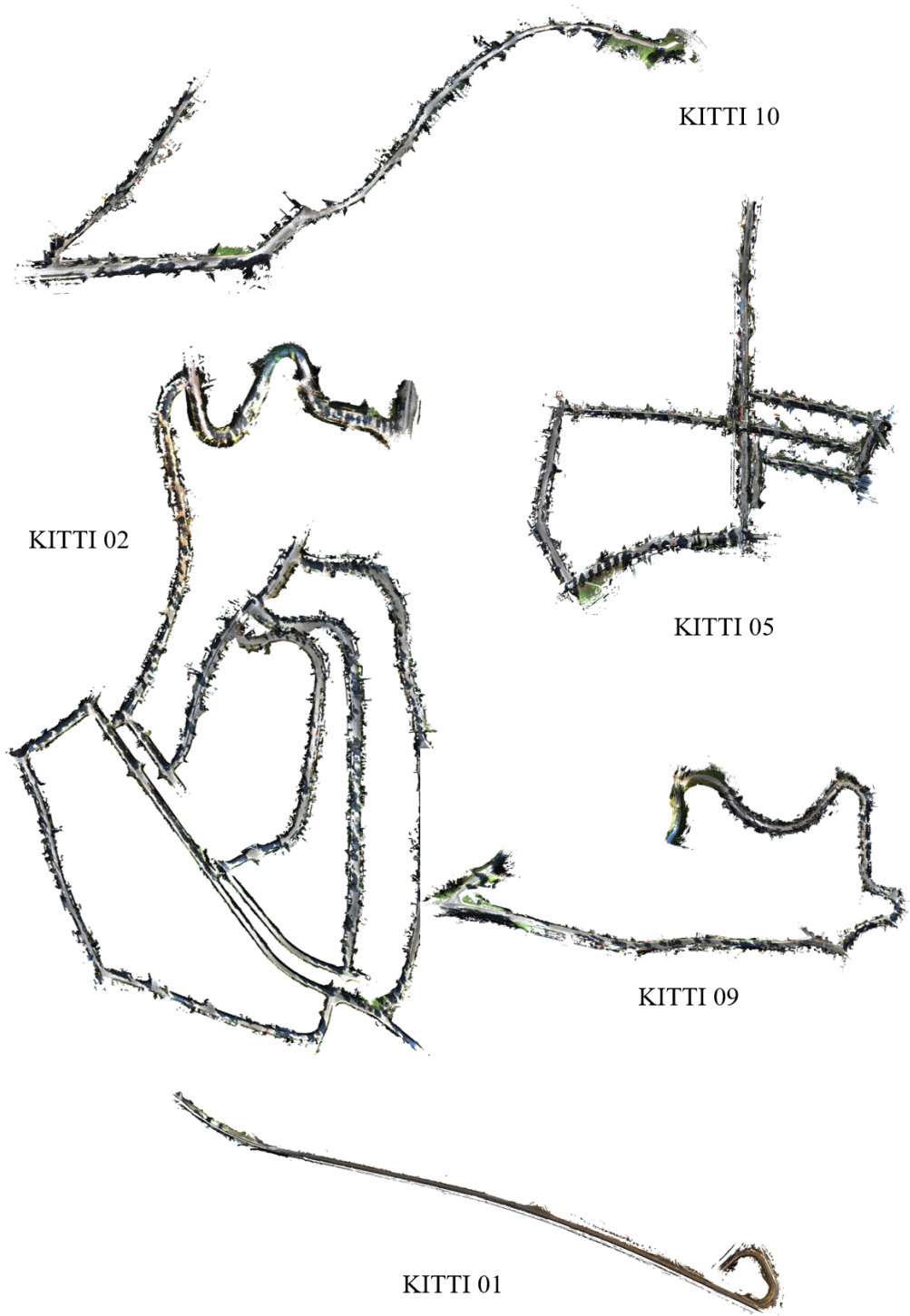}
    \caption{
        Qualitative reconstruction results on representative KITTI Odometry sequences. \ours{} maintains improved global consistency and reduces accumulated drift over long driving sequences.
    }
    \label{fig:kitti_appendix}
\end{figure*}

\cref{fig:kitti_appendix} visualizes estimated trajectories overlaid on the reconstructed scene for representative KITTI Odometry sequences, complementing the ATE-based analysis in Appendix~\ref{app:trajectory_visualization}.
These outdoor sequences contain long camera motions, limited image overlap, and repeated road structures.
By jointly optimizing high-frequency tracking factors and low-frequency wide-baseline factors, \ours{} maintains global trajectory consistency and reduces accumulated drift.

\end{document}